\documentclass[3p, preprint, 12pt]{elsarticle}

\usepackage{graphicx}
\usepackage{amssymb}

\usepackage{lineno}
\usepackage{threeparttable}
\usepackage{amsmath}
\usepackage{mathtools}
\usepackage{bm}
\usepackage{subcaption}
\usepackage{booktabs}
\usepackage{hyperref}
\hypersetup{colorlinks=true, linkcolor=blue!50!black, citecolor=blue!50!black, urlcolor=blue!50!black}
\usepackage{multirow}
\usepackage{xcolor}
\usepackage[ruled,vlined]{algorithm2e}
\usepackage{enumitem}
\usepackage{makecell}

\begin{document}

\begin{frontmatter}

\title{Transit-Oriented Land Use Planning Using Deep Reinforcement Learning}

\author[a]{Bo Yang}
\author[b]{Yuebing Liang\corref{cor1}}
\author[c]{Zhan Zhao}
\author[c]{Fangyi Ding}
\author[d]{Yunhan Zheng}
\author[e]{Shenhao Wang}
\author[f]{Jinhua Zhao}

\address[a]{Department of Civil and Environmental Engineering, University of California, Los Angeles, Los Angeles, CA 90095, USA}
\address[b]{Department of Urban Planning, Tsinghua University, Beijing 100084, China}
\address[c]{Department of Urban Planning and Design, The University of Hong Kong, Pokfulam Road, Hong Kong SAR, China}
\address[d]{College of Urban and Environmental Sciences, Peking University, Beijing 100871, China}
\address[e]{Department of Urban and Regional Planning, University of Florida, Gainesville, FL 32611, USA}
\address[f]{Department of Urban Studies and Planning, Massachusetts Institute of Technology, Cambridge, MA 02139, USA}

\cortext[cor1]{Corresponding author (liangyb@mail.tsinghua.edu.cn)}

\begin{abstract}
Transit-Oriented Development (TOD) is a key strategy for advancing sustainable urban development by promoting high-density, mixed-use land development within walking distance of public transit stations. However, traditional TOD planning often relies on heuristic methods, making it difficult to efficiently identify quantitatively optimized transit-oriented land use patterns. This paper presents DRL4TOD, a deep reinforcement learning framework that automatically optimizes land use types and floor area ratios (FARs) around transit stations in alignment with TOD principles. DRL4TOD incorporates a graph neural network to capture spatial interdependencies among land parcels, road networks, and metro stations, and uses a DRL-based optimization strategy to assign land-use types and FARs to selected parcels. It jointly optimizes multiple objectives, including land use density, land use diversity, and transit capacity utilization, while minimizing reconstruction costs to account for development feasibility. Applied to two real-world station areas in Shanghai, DRL4TOD consistently outperforms rule-based approaches and genetic algorithms, achieving more effective TOD outcomes with lower reconstruction costs. By adjusting objective weights, the framework enables flexible prioritization, such as placing greater emphasis on cost minimization when financial feasibility is a concern. These results demonstrate DRL4TOD's potential as a decision-support tool for generating coordinated land-use and FAR configurations under planner-specified priorities, rather than prescribing a single fixed design. For policymakers, the framework provides parcel-level, constraint-aware alternatives that make trade-offs among competing objectives explicit and support targeted decisions about development intensity and feasibility. By enabling alternatives to be generated and compared systematically alongside planner review, DRL4TOD can reduce reliance on repeated manual trial-and-error in TOD planning.

\end{abstract}


\end{frontmatter}

\section{Introduction} \label{sec:intro}

Urban land use planning plays a crucial role in shaping human mobility patterns and promoting sustainable transportation choices. However, traditional planning practices have often prioritized automobile-oriented development, with private vehicle use heavily influencing decisions on transportation infrastructure and land use \citep{okeke2020cities}. Over-dependence on private cars, however, has brought significant challenges to sustainable urban development, including traffic congestion, environmental degradation, socioeconomic disparities, and contributions to climate change \citep{abhijna2025new}.

In response to these challenges, Transit-Oriented Development (TOD) has emerged as a key strategy to reduce car dependency and promote sustainable urban development by promoting high-density, mixed-use development within walking distance of public transit. Initially conceptualized by \citet{calthorpe1993next}, TOD emphasizes the integration of land use and public transportation systems, structuring urban settlements around transit infrastructure. Theoretical frameworks underlying TOD posit that effective coordination of land use and transportation planning can reshape travel behavior, reduce reliance on private vehicles, and enhance the efficiency of public transit systems \citep{ewing2001travel,thomas2017defining}. Empirical studies substantiate these claims, demonstrating that TOD fosters sustainable travel patterns \citep{ibraeva2023impacts}, reduces greenhouse gas emissions\citep{ashik2022investigating}, revitalizes real estate markets \citep{su2021unraveling}, and promotes social cohesion \citep{zhao2024evaluating}.

To support TOD planning, researchers have developed quantitative frameworks for evaluating TOD performance around transit stations. The well-established 3Ds framework \citep{cervero1997travel}, comprising density, diversity, and design, provides a foundational approach for evaluating built environments in TOD contexts. Complementing this, the node-place model proposed by \citet{bertolini1999spatial} assesses transit supply and land-use characteristics. Subsequent extensions to these frameworks incorporate additional metrics, such as transit capacity utilization and development feasibility, to capture the multidimensional nature of TOD \citep{reusser2008classifying, vale2015transit, singh2017measuring, cucuzzella2022tod}. While these frameworks have significantly advanced the understanding of TOD outcomes, they are predominantly descriptive and evaluative, providing limited support for answering a more actionable planning question: how should land use configurations within station areas be adjusted to better achieve TOD objectives?

Beyond evaluation, recent studies have developed optimization-based methods to generate TOD-aligned land-use configurations \citep{wang2026multi}. These studies typically represent station areas as raster grids and employ heuristic optimization methods, such as genetic algorithms (GA), to allocate land-use types and floor area ratio (FAR) values across vacant grids \citep{li2010transit, ma2018sustainable, sahu2018methodology, ding2022integrating}. While these methods demonstrate the potential of quantitative optimization for TOD planning, they leave related gaps in station-area planning. First, raster-based representations do not preserve the irregular parcel structure and explicit parcel--road relationships of real urban areas. Second, land-use type and FAR are spatially interdependent planning decisions: the assignment at one parcel can alter the context of neighboring parcels and affect aggregate outcomes such as transit capacity utilization, land-use diversity, and development feasibility. As the number of parcels and candidate land-use/FAR combinations grows, this interdependence causes the joint decision space to expand combinatorially, and conventional heuristic methods---which typically evaluate the fitness of a complete candidate solution as a whole rather than how individual, spatially adjacent decisions jointly produce that outcome---become increasingly difficult to search efficiently as problem dimensionality increases. A suitable method must therefore combine parcel-level spatial representation with explicit modeling of spatial interdependence among planning decisions. Graph-based deep reinforcement learning offers a way to address these requirements. A spatial graph can represent metro stations, roads, and land parcels as connected entities, and GNN message passing updates each parcel's representation using information from neighboring nodes and edges, in contrast to the gene-based encodings used in heuristic search methods. The resulting state representation allows a DRL policy to condition each land-use and FAR decision on the current neighborhood and station context, enabling the joint optimization of spatially interdependent planning decisions more directly than existing heuristic approaches.

Reflecting this potential, there has been growing interest in applying AI more broadly to urban planning decisions \citep{hosseini2025towards}. A number of studies have explored DRL-based approaches for various transportation planning tasks, including metro network expansion \citep{su2024metrognn}, road layout design in informal settlements \citep{zheng2023road}, and electric bus network planning \citep{zhao2025large}. In the realm of land use planning, \citet{zheng2023spatial} integrated DRL with graph neural networks (GNNs) to optimize land use allocation plans; however, this approach considers only land-use type and does not model development intensity (FAR) or the reconstruction costs associated with redeveloping existing built environments, both central to TOD planning in built-up urban contexts. As a result, despite growing evidence that graph-based DRL can capture spatial interdependence in related urban planning problems, its application to TOD planning specifically remains unexplored.

To address these challenges, this paper introduces DRL4TOD, a deep learning framework designed to support TOD planning by automatically optimizing land use configurations around transit stations. The urban environment is modeled as a graph, capturing spatial relationships among stations, surrounding land parcels, and road networks. A graph learning model encodes these relationships, and the embedded features are used by a DRL algorithm to identify optimal design strategies. Empirical validation is done based on two real-world cases in Shanghai. The specific contributions of this study are summarized as follows:

\begin{itemize}[noitemsep]
    \item We develop a planner-guided DRL framework for transit-oriented land-use planning that automatically generates and evaluates alternative spatial configurations under planner-defined objectives, constraints, and priorities.
    \item We formulate TOD planning as a graph-based spatial decision problem, representing parcels, roads, and transit stations within a unified spatial structure and jointly optimizing parcel-level land-use types and FARs.
    \item We design a multi-objective reward function that balances density, diversity, transit-capacity utilization, and reconstruction cost, enabling planners to explore trade-offs between TOD performance and implementation feasibility.
    \item Using two station areas in Shanghai, we show that DRL4TOD outperforms a variety of baselines, achieving higher TOD performance with lower reconstruction costs.
\end{itemize}

\section{Literature Review}
\label{sec:literature}

\subsection{Transit-Oriented Land Use Planning}
\label{subsec:tod_related_research}

A substantial body of empirical research has established the role of TOD in advancing sustainable urban development. TOD neighborhoods have been shown to reduce automobile dependency, promote transit-oriented commuting, and encourage active travel \citep{ewing2001travel,thomas2017defining,xiao2021subway,ibraeva2023impacts}. At the urban scale, TOD has also been associated with lower transport-related carbon emissions \citep{ashik2022investigating,abhijna2025new}, while more recent studies have examined its potential to improve accessibility and quality of life for lower-income residents \citep{zhao2024evaluating}. Collectively, these findings highlight TOD's potential to advance mobility efficiency, environmental sustainability, and social equity.

Accordingly, a substantial literature has focused on how station areas should be evaluated and planned. One major strand of research develops indicators for assessing TOD performance. A foundational framework in this domain is the 3D framework—density, diversity, and design—proposed by \citet{cervero1997travel}, which was subsequently extended to the 5D framework by incorporating destination accessibility and distance to transit \citep{ewing2010travel}. Complementing these built-environment dimensions, the node-place model characterizes station areas through two dimensions: their accessibility and connectivity as transport nodes, and the intensity and diversity of surrounding urban activities as places \citep{bertolini1999spatial}. Subsequent studies have further expanded TOD evaluation to include operational measures such as transit-capacity utilization \citep{singh2017measuring,zhou2020using} and implementation considerations such as land availability, demolition cost, construction feasibility, and regulatory constraints \citep{cucuzzella2022tod,mangu2025evaluation}. These studies provide increasingly comprehensive criteria for assessing station-area performance, but they primarily evaluate existing conditions or predefined alternatives rather than generate new spatial configurations.

A second strand of research focuses on plan generation and optimization. Multicriteria decision-analysis (MCDA) approaches provide structured means of ranking predefined planning alternatives according to multiple criteria and stakeholder preferences \citep{banai1998transit,kunz2010analytic}, but their outcomes remain dependent on the alternatives specified in advance. In parallel, optimization-based approaches directly search for new spatial configurations. \citet{lin2006tod} developed an early TOD optimization model to determine floor-area ratios for different land uses around transit stations, considering objectives including transit ridership, living environment, and social equity. Subsequent studies applied genetic algorithms to jointly optimize land-use types and development densities within transit catchment areas, often with the aim of increasing transit use \citep{li2010transit,sahu2018methodology,ma2018sustainable}. More recently, \citet{ding2022integrating} extended this line of research by jointly optimizing land use and the greenway network around metro stations using an improved immune genetic algorithm.

Despite this progress, two limitations remain important for operational TOD planning. First, many optimization studies represent station areas using regular grids, whereas planning decisions are implemented on irregular parcels with heterogeneous sizes, shapes, existing development conditions, and regulatory constraints. This mismatch limits the direct applicability of generated solutions to real planning practice. Second, existing methods provide only limited representation of the spatial interdependence among parcel-level land-use choices, development intensity, road structure, and transit accessibility. Yet these decisions are inherently coupled: the suitability and performance of one parcel depend on both neighboring development and its relationship to the surrounding transport network. These limitations point to the need for a parcel-based optimization framework that explicitly represents spatial relationships while jointly considering land-use allocation, development intensity, and planning feasibility.

\subsection{Deep Reinforcement Learning for Urban Planning}

Deep reinforcement learning (DRL) is well suited to urban planning problems in which decisions are sequential, the action space is large, and each action alters the state of the environment \citep{sutton2018reinforcement}. These characteristics are common in spatial planning, where the location and intensity of one intervention influence subsequent decisions and the performance of the resulting configuration. Combined with graph neural networks (GNNs), DRL can further represent spatial relationships among heterogeneous urban elements rather than treating planning units as independent entities.

Recent studies have applied DRL and graph-based learning to several urban planning and transportation problems. \citet{su2024metrognn} developed a graph-based RL framework for metro network expansion; \citet{liang2024sponet} formulated facility-location planning as a Markov decision process solved with DRL; and \citet{zhao2025large} applied heterogeneous graph-based DRL to the joint planning of electric bus fleets and charging facilities. These studies demonstrate DRL's capacity to handle structured spatial environments and sequential decisions, but their planning tasks concern transportation infrastructure or facility placement rather than the integrated land-use and transit objectives central to TOD. In land-use planning specifically, \citet{zheng2023spatial} developed a DRL model that generates spatial plans to improve access to services and green space, demonstrating DRL's potential for spatial-configuration generation. However, this model addresses land-use placement alone: it does not jointly assign land-use type and development intensity using FAR, incorporate a transit-capacity objective, or account for development feasibility in built-up station areas. Existing DRL-based land-use studies therefore have yet to directly address the specific requirements of TOD planning.

\section{Preliminaries}
\label{sec:preliminaries}

\subsection{Deep Reinforcement Learning}
\label{preliminary:DRL}

This subsection provides a brief introduction to deep reinforcement learning (DRL), a technique well-suited for complex sequential decision-making tasks and central to the methodology of this study. In a typical DRL framework, an agent interacts with the environment in discrete time steps: it observes the current state, selects an action, receives a reward, and transitions to a new state. Through repeated interactions, the agent learns a policy that maximizes cumulative rewards over time. The integration of deep learning allows DRL to extract hierarchical features and model nonlinear relationships, making it particularly effective for complex decision-making problems.

The underlying decision process in DRL is formalized as a Markov Decision Process (MDP), which provides a principled mathematical framework for sequential decision-making under uncertainty \citep{bertsekas2012dynamic}. An MDP is defined by a 4-tuple $M=(S, A, P_a, R_a)$, where $S$ denotes the set of possible states, $A$ the set of available actions, $P_a(s, s')$ the transition probability from state $s$ to $s'$ under action $a$, and $R_a(s, s')$ the immediate reward received after the transition. The Markov property assumes that future states depend only on the current state and action, ensuring temporal consistency and tractability.

To solve an MDP, one of the most widely used DRL algorithms is Proximal Policy Optimization (PPO), which achieves a balance between exploration and policy stability. PPO employs two neural networks: a policy network that maps states to a distribution over actions, and a value network that estimates the expected return of each state. The policy network is trained by maximizing a clipped surrogate objective that constrains the step size of policy updates, thereby avoiding destructive policy shifts. Concurrently, the value network minimizes the mean squared error between predicted and actual returns. This coordinated learning strategy enables stable and sample-efficient training, making PPO particularly suitable for large, continuous, or structured action spaces encountered in real-world planning problems.

\subsection{Problem Definition}
\label{sec:problem_definition}

TOD planning aims to optimize the spatial organization of land parcels around metro stations to improve both transportation efficiency and urban livability. In this study, the initial state is defined by the current conditions of the station area, including the location of the transit station, the existing distribution of land-use types, and the layout of the road network. In addition to these physical site conditions, the model also incorporates planning needs and requirements as input constraints. Planning needs refer to the target land use composition specified by planners, i.e., the desired amount/area for each land use type within the study area. Planning requirements refer to regulatory constraints, such as the minimum parcel size and the allowable FAR ranges for each land use category.

The planning task involves jointly optimizing land use allocation and the corresponding FAR values of parcels around metro stations. The optimization objective is twofold: (1) to maximize TOD-related outcomes, such as increased residential, business and office densities, greater land use diversity, and improved transit capacity utilization; and (2) to minimize the scale of reconstruction, thereby reducing the associated financial burden. By integrating multiple, and often competing, planning objectives, the proposed formulation captures the inherent complexity of real-world TOD projects and improves the feasibility of the resulting planning solutions. The specific land-use categories and FAR limits used to instantiate these planning needs and requirements in Shanghai are reported in Section~\ref{sec:experiment_design}.

\subsection{Planner-Guided Decision-Support Workflow}
\label{sec:planning_workflow}

\begin{figure}[!htb]
  \centering
  \includegraphics[width=1.0\textwidth]{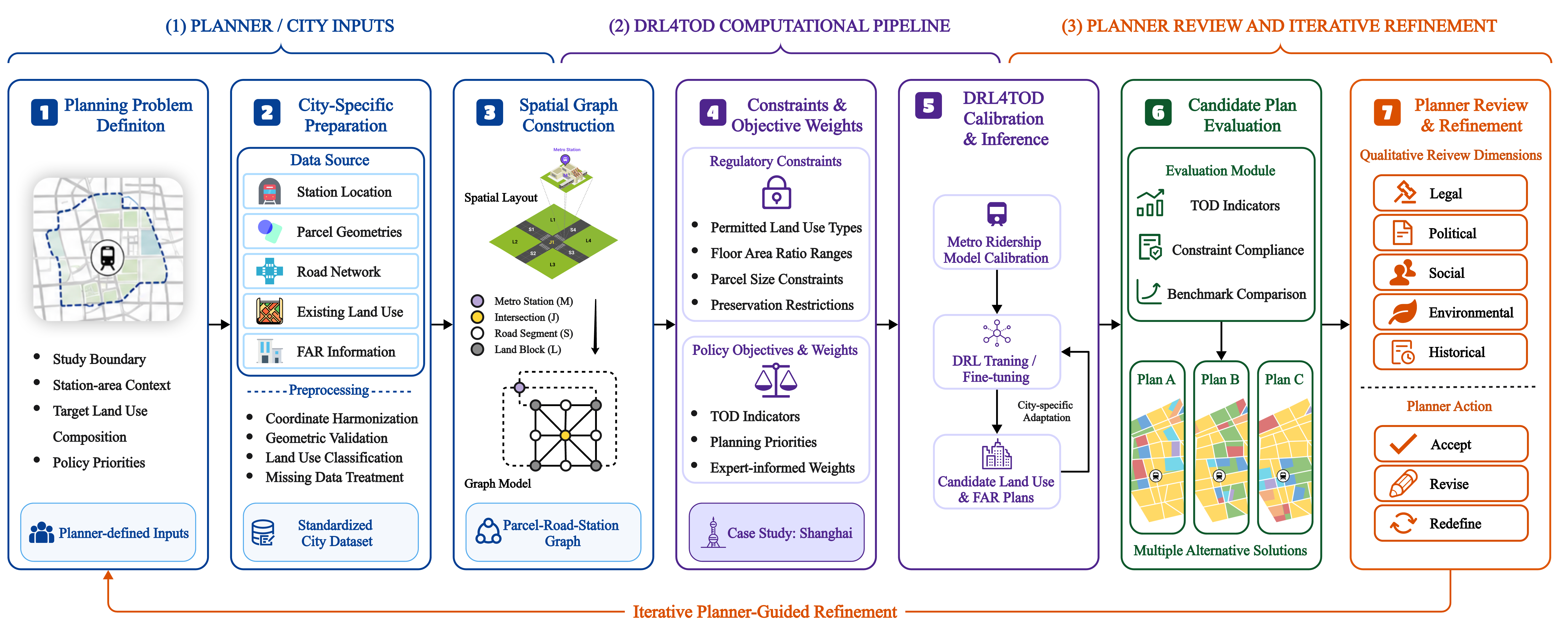}
  \caption{Planner-Guided Decision-Support Workflow}
  \label{fig:planning_workflow}
\end{figure}

DRL4TOD is designed as a planner-guided decision-support pipeline rather than an autonomous planning system. The planner defines the planning problem, specifies regulatory constraints and policy priorities, and reviews the resulting alternatives, while DRL4TOD translates these inputs into a spatial optimization problem, generates candidate land-use and FAR configurations, and evaluates their quantitative performance. Figure~\ref{fig:planning_workflow} summarizes this division of roles across the workflow.

The workflow begins with planner-led problem definition and data preparation (Figure~\ref{fig:planning_workflow}(1)--(2)). The planner specifies the station-area boundary, target land-use composition, regulatory requirements, and other local planning needs described in Section~\ref{sec:problem_definition}. The required spatial inputs -- including station locations, parcel geometries, road networks, existing land uses, and FAR information -- do not need to follow a standardized data format, as long as they can be processed into the spatial elements required by DRL4TOD. Official cadastral or planning-authority data are preferred where available because they are generally more accurate and current. Otherwise, these inputs can be derived from widely available open geospatial sources, including OpenStreetMap \citep{OpenStreetMap}, Microsoft Building Footprints \citep{MicrosoftBuildingFootprints}, Google Open Buildings \citep{GoogleOpenBuildings}, and remote-sensing-based extraction methods. Building footprint and height information can further be used to estimate development intensity where official FAR records are unavailable.

DRL4TOD then converts these inputs into the parcel--road--station graph used for optimization (Figure~\ref{fig:planning_workflow}(3)). Planner-specified regulatory requirements are encoded as constraints on permitted land uses, FAR ranges, and parcel eligibility, while policy priorities are represented through the relative objective weights defined in Section~\ref{method:reward} (Figure~\ref{fig:planning_workflow}(4)). The model is subsequently calibrated and trained for the local context and generates candidate land-use and FAR plans, together with their performance on the specified TOD objectives and planning constraints (Figure~\ref{fig:planning_workflow}(5)--(6)). For the Shanghai application, the baseline objective weights were informed by the literature, and an alternative planner-defined weighting scheme is examined in Section~\ref{subsec:workflow_case_demo}.

The final decision remains with the planner (Figure~\ref{fig:planning_workflow}(7)). Planners interpret the model-generated alternatives in light of legal, political, social, environmental, infrastructure, and other considerations that are not fully represented by the quantitative objectives. They may therefore accept a candidate, modify it, revise the constraints or objective weights, or redefine the planning problem and rerun DRL4TOD. In this way, the model supports iterative exploration of planning alternatives while leaving interpretation, validation, and final judgment to the planner.

\section{Methodology}
\label{sec:methodology}

This section presents DRL4TOD as an integrated, planner-guided decision-support framework that uses automated optimization to generate and evaluate candidate land-use and FAR configurations around transit stations in alignment with TOD principles. The framework connects a planning application workflow, in which local planning needs, regulatory requirements, and policy priorities are specified and generated alternatives are reviewed, with a graph-based deep reinforcement learning model that represents spatial relationships among parcels, roads, and stations and sequentially allocates the specified land-use types by selecting feasible parcels and assigning FAR values.

The section is organized according to this progression from planning inputs to model architecture. Section~\ref{sec:problem_definition} defines the planning problem, including the planning needs and regulatory requirements introduced by the workflow, while Section~\ref{sec:planning_workflow} describes the workflow for configuring and applying DRL4TOD to a station area and adapting it to other urban contexts. Section~\ref{sec:framework_overview} summarizes the computational components, and Section~\ref{sec:graph_representation} details the parcel--road--station graph on which these components operate. The remaining subsections formalize the model itself: the MDP formulation (Section~\ref{method:mdp}), graph-based state encoder (Section~\ref{sec:graph_state_encoder}), policy and value networks (Sections~\ref{sec:policy_function} and~\ref{sec:value_network}), and reward function (Section~\ref{method:reward}).

\subsection{DRL4TOD Framework Overview}
\label{sec:framework_overview}

\begin{figure}[htbp]
  \centering
  \includegraphics[width=1\textwidth]{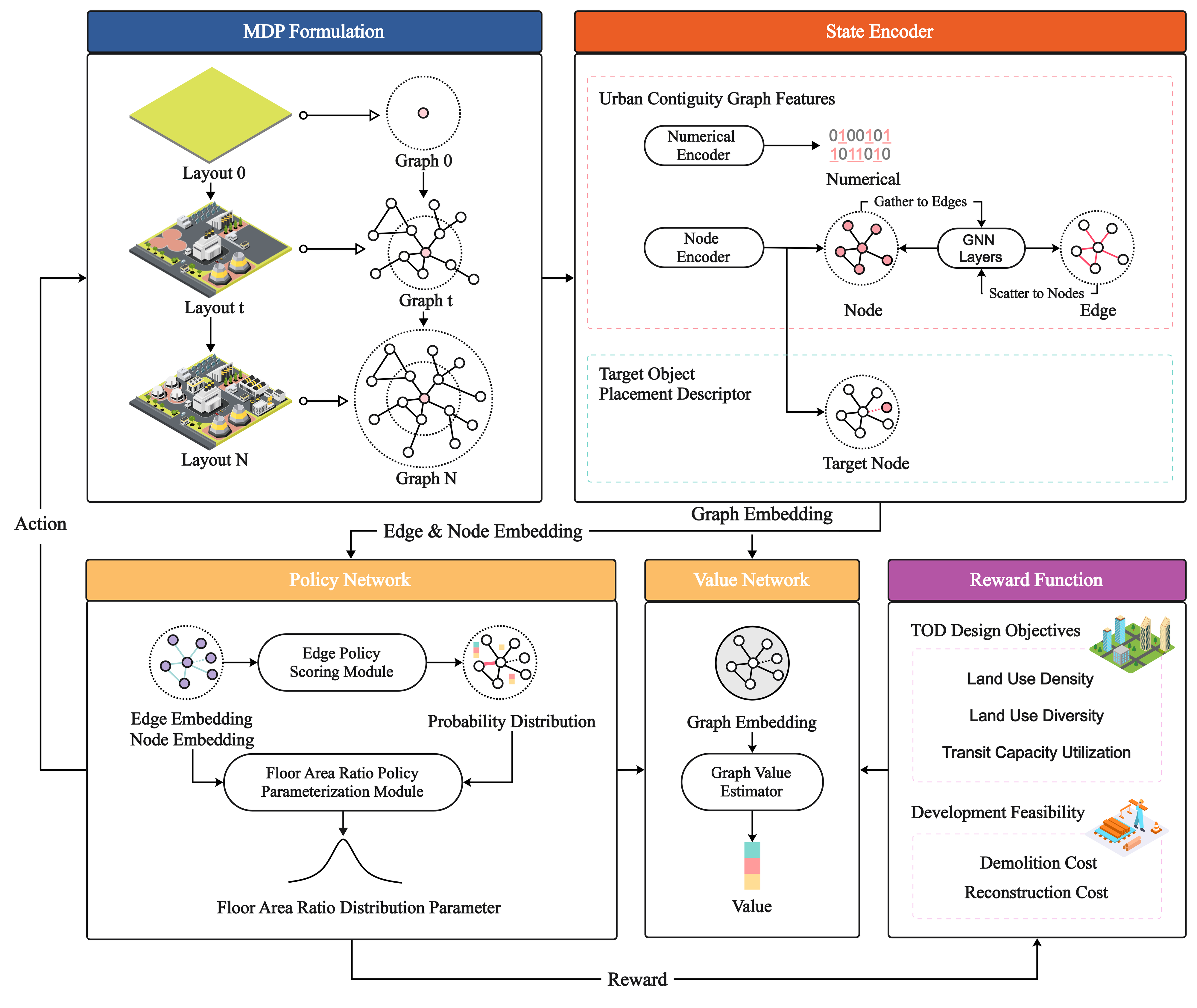}
  \caption{Overview of the proposed DRL4TOD framework}
  \label{fig:Methodology}
\end{figure}

As illustrated in Figure~\ref{fig:Methodology}, DRL4TOD comprises five core components: (1) MDP Formulation, formulating the TOD planning problem as an MDP by modeling the urban layout surrounding metro stations as a dynamic graph (see Section~\ref{method:mdp}); (2) State Encoder, utilizing a graph neural network to encode the current urban configuration into latent feature representations (see Section~\ref{sec:graph_state_encoder}); (3) Policy Network, which selects planning actions based on encoded state representations (see Section~\ref{sec:policy_function}); (4) Value Network, estimating the expected returns of each state to guide policy optimization (see Section~\ref{sec:value_network}); (5) Reward Function, evaluating the outcomes of planning actions by integrating TOD objectives and development feasibility (see Section~\ref{method:reward}).

The policy network and value network perform complementary roles. The policy network is the decision component of the agent: it produces probabilities for feasible planning locations and determines the FAR assigned to the selected location. The value network is the evaluation component: it estimates the expected cumulative reward associated with the current planning state and provides a reference for updating the policy during PPO training. Thus, the policy network generates planning actions, whereas the value network supports the learning process and does not directly generate the final plan.

The DRL4TOD framework operates by iteratively evolving the urban spatial configuration through sequential planning decisions. At each decision step, the agent constructs a graph-based representation of the current urban form and encodes it into latent features via the state encoder. The policy network then selects an action, comprising the choice of a location to modify and the corresponding FAR adjustment. The urban layout is updated based on this action. Subsequently, the value network estimates the utility of the new state, while the reward function provides immediate feedback reflecting TOD-aligned objectives. This reward signal, combined with value estimations, is used to train both the policy and value networks using the PPO algorithm, promoting stable and efficient convergence toward TOD-compatible land use configurations.

\subsection{Representing Urban Spatial Layout as a Graph}
\label{sec:graph_representation}

Traditional raster-based spatial data models have been widely employed in transit-oriented land use optimization \citep{ma2018sustainable, gao2021sustainable, li2018improved}. These models discretize the spatial domain into uniform grids, with each cell assigned a specific land-use category. While raster representations facilitate spatial analysis through their simplicity, they inadequately reflect real-world urban forms, where land parcels are typically irregular polygons structured around transportation infrastructure.

To address these challenges, we adopt a vector-based spatial model following \citet{zheng2023spatial}, wherein urban elements are represented using more accurate geometric primitives: points, lines, and polygons. In this representation, \textit{points} denote discrete elements (e.g., metro stations, road intersections), \textit{lines} represent linear elements (e.g., road segments, parcel boundaries), and \textit{polygons} define land parcels corresponding to urban blocks.

Building upon the vector representation, we construct a graph-based model to capture spatial relationships among urban elements. The urban layout is formalized as a graph $G = (V, E)$, where the node set $V$ comprises three types:
\begin{itemize}
    \item \textit{Point Nodes}: Represent metro stations and intersections. Each point node is associated with a feature vector that encodes its type and centroid coordinates. For metro station nodes, which serve as anchors for TOD, an 800-meter walking radius defines their service buffer \citep{cervero2007transit}.
    \item \textit{Line Nodes}: Represent road segments and parcel boundaries. Each line node has a feature vector indicating its type, centroid coordinates, and geometric attributes, including length as well as the width and height of its bounding box.
    \item \textit{Polygon Nodes}: Represent land parcels.  Each land parcel is described by features comprising: (1) functional attributes (land use type and FAR); (2) locational attributes (centroid coordinates); and (3) geometric and morphological attributes as defined in \citet{zheng2023spatial}, including area, perimeter, the width and height of its bounding box, and shape regularity metrics such as rectangularity, equivalent rectangle index, and compactness.
\end{itemize}

The edges $E$ capture spatial relationships based on geographic proximity and topological adjacency. Specifically, the metro station node is connected to land parcels within an 800-meter walking radius, reflecting their service area. Land parcels are connected to adjacent parcels to represent physical contiguity. Road segments are linked to intersections and neighboring segments, forming a continuous road network. Intersections are similarly connected to nearby road segments and, where applicable, to other intersections. Together, these connections encode the spatial structure of the built environment surrounding the metro station.

\begin{figure}[!htbp]
  \centering
  \includegraphics[width=0.8\textwidth]{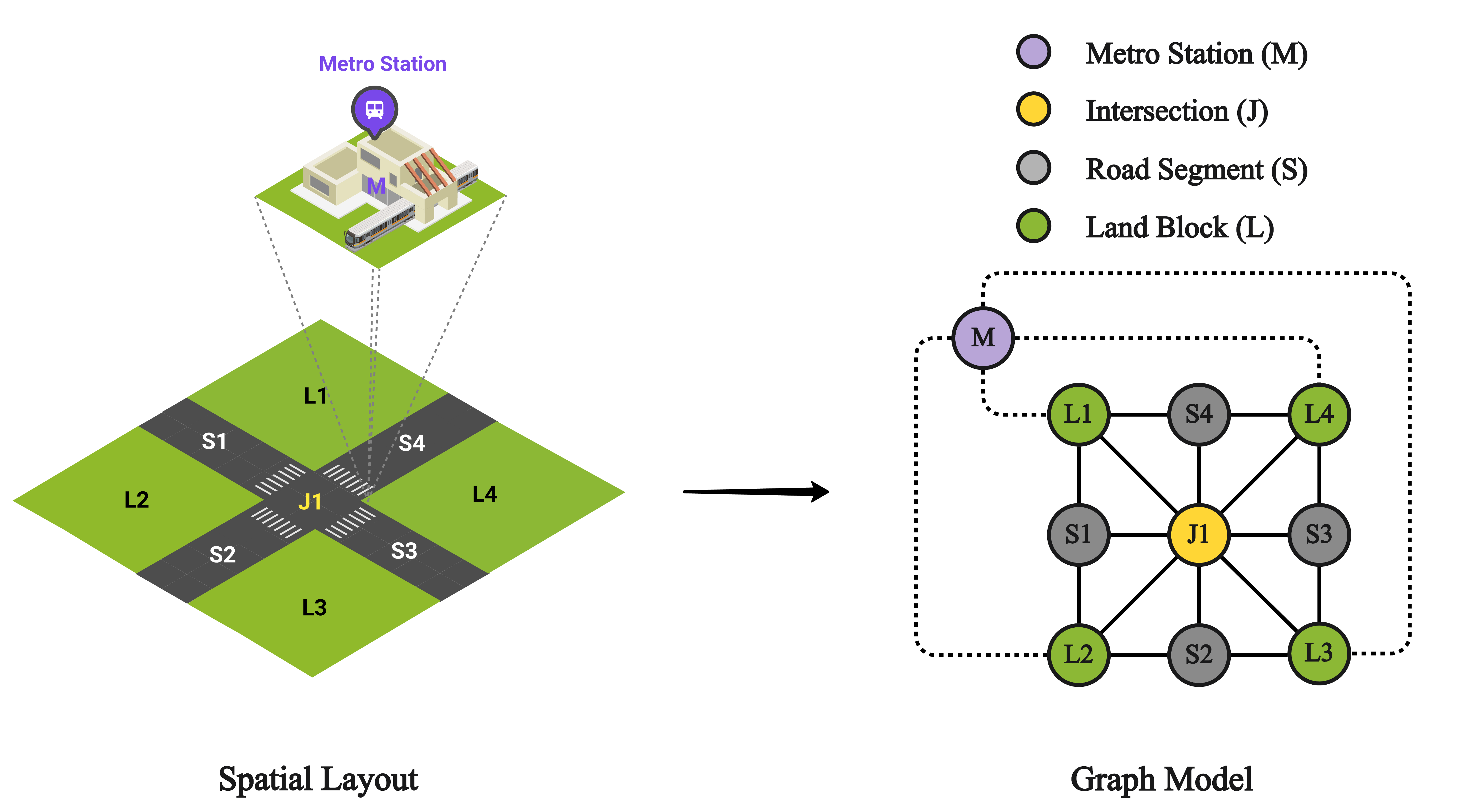}
  \caption{Graph construction from spatial layout in the TOD planning context}\label{fig:graph_construction}
\end{figure}

Through this formulation, the graph $G$ encodes both the geometric structure and the functional relationships of the urban environment, providing a hybrid spatial-topological framework suitable for modeling and analysis in TOD contexts. Figure~\ref{fig:graph_construction} illustrates an example of how an urban spatial layout is transformed into a graph representation in an intuitive manner.

\subsection{Formulating TOD Planning as an MDP}\label{method:mdp}

As introduced in Section~\ref{preliminary:DRL}, an MDP provides a robust mathematical foundation for modeling sequential decision-making tasks within DRL frameworks. In this subsection, we formulate the TOD planning problem as an MDP. In DRL4TOD, the TOD planning principles are operationalized through the state, action, transition, and reward components of the MDP. The state describes the current station-area configuration, including parcel, road, station, land-use, and FAR attributes. The action selects a feasible planning location and assigns the corresponding FAR for the target land-use type, while the transition updates the spatial configuration after each decision. Density and land-use diversity are evaluated directly through the LUDE and LUDI reward components, respectively. Distance to transit is represented through the station service buffer and the distance-decay weighting in the LUDE metric, while destination accessibility is captured through the transit-capacity-utilization component, which estimates ridership from surrounding built-environment features. Development feasibility is included as an additional reward component. Spatial design -- block-scale and pedestrian-network qualities such as street connectivity and building frontage -- is not represented as an explicit reward objective in the current formulation, but is captured only indirectly through parcel-level morphological features.

Building on the spatial graph representation (Section~\ref{sec:graph_representation}), the urban environment around each metro station is modeled as a dynamic graph, where nodes represent urban elements and edges capture spatial proximity. The initial MDP state encodes the existing built environment. Following \cite{zheng2023spatial}, we predefine the sequence of land use types to be placed. At each step, the agent selects a parcel for the given land use type and assigns its floor area ratio, thereby updating the spatial configuration and the corresponding graph-based state.

In this context, the MDP is defined as $M=(S, A, P_a, R_a)$, where each component is specified as follows:

\begin{itemize}
    \item \textbf{States (\(S\)):} Each state \( s_t \) encapsulates the spatial and functional status of the planning area at time step \( t \), providing the agent with essential context for decision-making. It comprises two elements: (1) spatial graph representation \( s_{\text{graph}} \), which models the environment as a graph \( G_t = (V_t, E_t) \), where nodes \( v_t \in V_t \) represent urban elements with spatial and functional attributes, and edges \( e_t \in E_t \) capture spatial proximity at time step $t$; (2) planning progress summary \( s_{\text{progress}} \), which tracks the development status of each land use type. Specifically, it includes the current area proportions, number and FAR values of assigned parcels across different land use types. Together, these components provide the agent with a comprehensive view of the current planning context; and (3) the target land use \( s_{\text{target}} \), which encodes the geometric and functional requirements of the next land use type to be placed.
    \item \textbf{Actions (\(A\)):} At each step, given the target land use type, the agent selects an action \( a_t \in A \) consisting of: (1) location selection, where the agent either selects an existing land parcel or subdivides it to create a new sub-parcel based on the functional requirements of the land use type; and (2) FAR assignment, where a floor area ratio is allocated to define the development intensity of the selected or newly created parcel. %
    \item \textbf{Transitions (\(P_a\)):} The environment evolves deterministically in response to the agent’s actions. If the agent selects an existing parcel, its attributes are updated to reflect the assigned land use and FAR. If the agent subdivides a parcel, a new node is added to represent the sub-parcel, while the attributes of the original parcel and the connectivity among urban elements—such as roads, intersections, and metro stations—are updated to reflect the modified spatial structure.
    \item \textbf{Reward ($R_a$):} The reward provides feedback on the desirability of a planning decision, guiding the agent toward goal-aligned outcomes. Intermediate rewards are set to zero during the planning process, with a final reward computed upon completion. This final reward is formulated as a weighted sum of multiple evaluation objectives, which reflects TOD priorities and development feasibility, ensuring that both planning quality and practical constraints are considered.
\end{itemize}
This MDP formulation enables the integration of spatial interdependencies, planning rules, and long-term objectives within a reinforcement learning framework, facilitating iterative improvement through agent–environment interactions.

\begin{figure}[!htbp]
  \centering
  \includegraphics[width=1\textwidth]{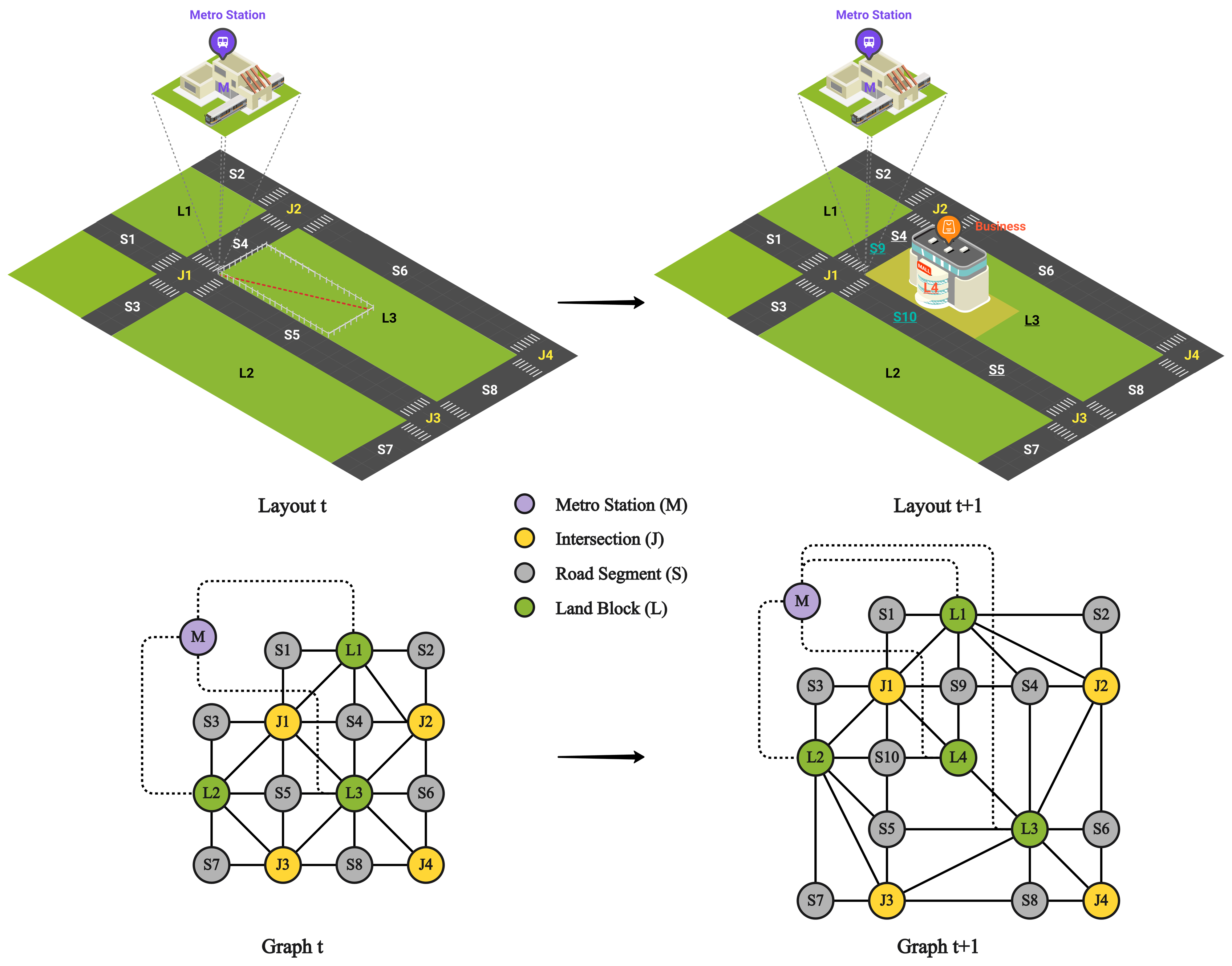}
  \caption{Illustration of MDP transitions with spatial and graph representations}
  \label{fig:mdp_state_transition}
\end{figure}

Figure~\ref{fig:mdp_state_transition} illustrates how the urban layout and its corresponding spatial graph evolve across two consecutive decision steps \( t \) and \( t{+}1 \) in the MDP framework. At each time step, the agent selects a planning action that modifies the current urban layout—specifically, it chooses a development location and assigns a corresponding FAR value. The location is determined by selecting an edge that connects an intersection node (denoted as \( J \)) to a land parcel node (denoted as \( L \)), representing a feasible planning candidate.

\subsection{Graph-Based State Encoder}
\label{sec:graph_state_encoder}

To effectively represent the dynamic and spatially structured TOD planning environment, we design a graph-based state encoder that integrates three inputs introduced in Section~\ref{sec:graph_representation} into a unified latent representation: (1) the spatial graph \(G=(V, E)\), where nodes represent urban parcels with spatial-functional attributes and edges reflect spatial proximity; (2) the planning progress summary \( S_{\text{progress}} \), which encodes the current area proportion, number of parcels, and FAR values of planned land uses; and (3) the target land use type to be placed, modeled as a virtual node $v_{target}$ to capture its interaction with the existing spatial layout.

The target virtual node represents the land-use type that is scheduled to be placed at the current decision step. For example, if the next target is office use, the virtual node encodes "office" as the current planning requirement. Unlike a parcel node, it does not represent a physical location, parcel geometry, or existing development, has no edges connecting it to the spatial graph, and is not itself selected as an action. Instead, it provides a semantic reference that is refined using the representations of physical parcels, roads, intersections, and metro stations after these have incorporated their own spatial context through message passing. This allows the model to assess how the pending land-use type relates to its surrounding spatial context before the policy network selects a feasible parcel and assigns its FAR.

These inputs are processed through the following three steps: in the first step, input features are encoded into latent representations. Each node \( v_i \in V\) and the target virtual node \( v_{\text{target}} \) are processed through a shared node encoder \( f_{\text{node}} \), composed of fully connected layers, to generate their respective embeddings \( \mathbf{v}_i \) and \( \mathbf{v}_{\text{target}} \). Meanwhile, the planning progress summary \( s_{\text{progress}} \), represented as a numerical vector, is encoded using a dedicated feature encoder \( f_{\text{progress}} \), producing the latent representation \( \mathbf{s}_{\text{progress}} \).

In the second step, a graph neural network is applied to capture spatial dependencies and contextual relationships among urban elements. At each GNN layer \( k \), edge embeddings  \( \mathbf{e} \) are computed based on the features of adjacent nodes, and node embeddings are subsequently updated by aggregating messages from their neighbors:
\begin{gather}
    \mathbf{e}_{ij}^{(k)} = \frac{1}{2} \left[ \tanh\left( \text{fc}^{(k)}(\mathbf{v}_i^{(k-1)} \Vert \mathbf{v}_j^{(k-1)}) \right) + \tanh\left( \text{fc}^{(k)}(\mathbf{v}_j^{(k-1)} \Vert \mathbf{v}_i^{(k-1)}) \right) \right], \\
    \mathbf{v}_i^{(k)} = \mathbf{v}_i^{(k-1)} + \frac{1}{|\mathcal{N}_i| + \epsilon} \sum_{j \in \mathcal{N}_i} \mathbf{e}_{ij}^{(k)}
\end{gather}
where \( \mathbf{v}_i^{(k)} \) denotes the embedding of node \( v_i \) at layer \( k \), \( \text{fc}^{(k)} \) is a layer-specific fully connected transformation, \( \mathcal{N}_i \) is the neighbor set of \( v_i \), and \( \epsilon \) is a small constant added for numerical stability. For a parcel node, this update combines its own attributes with learned messages from spatially adjacent parcels, roads, intersections, and, where connected through the graph, metro-station nodes. Repeated message-passing layers allow the representation to incorporate multi-hop neighborhood and station context. The resulting embedding can therefore represent spatially conditioned relationships among land uses and accessibility features, including potential complementarities among uses located near a transit station when such configurations improve the planning objectives. The policy network uses these context-aware representations when selecting the next parcel and assigning its FAR, thereby enabling sequential decisions that respond to the evolving spatial configuration.

In the third step, we model the interaction between the spatial layout and the target land use type \( v_{\text{target}} \) using a self-attention mechanism. This mechanism refines the target embedding \( \mathbf{v}_{\text{target}} \) by conditioning it on the node embeddings of all graph nodes:
\begin{equation}
    \mathbf{v}_{target} = \sum_{v_i \in V} w_i \mathbf{v}_{i},
\end{equation}
where $w_i$ represents the adaptive attention weights computed to quantify the influence of each node \( v_i \in V \) on the target, allowing \( \mathbf{v}_{\text{target}} \) to selectively integrate relevant spatial context. This enriched target embedding subsequently serves as a core component of the input to the policy and value networks, as detailed in Section~\ref{sec:policy_function}. %

\subsection{Policy Network}
\label{sec:policy_function}

Building on the encoded components described in Section~\ref{sec:graph_state_encoder}, we construct the input to the policy network, denoted as \( \mathbf{x}_{\pi} \), by concatenating the edge embeddings \( \mathbf{e} \) with the target-aware land use embedding \( \mathbf{v}_{\text{target}} \). This joint representation captures both localized spatial structure and planning intent, providing rich context for action selection.

As introduced in Section~\ref{method:mdp}, the TOD planning agent performs two types of actions: selecting a candidate location and assigning a corresponding FAR value. The location is chosen by selecting an edge that connects an intersection to a developable parcel, while the FAR determines the development intensity for that location. Accordingly, the policy network is designed as a multi-task framework that simultaneously handles discrete edge selection and continuous FAR assignment.

For edge selection, the policy network models a categorical distribution over feasible edges. To ensure valid decisions, a feasibility mask is maintained to exclude edges connected to already-assigned parcels. Given the policy input \( \mathbf{x}_{\pi} \), an edge scoring module \( f_{\text{edge}} \) composed of fully connected layers produces unnormalized logits \( \mathbf{z} \in \mathbb{R}^{N_{\text{edge}}} \). These logits are masked and normalized via a softmax function to yield the edge selection probabilities:
\begin{equation}
    \mathbf{z} = f_{\text{edge}}(\mathbf{x}_{\pi}), \quad
    \pi_{\theta}^{\text{edge}}(a^{\text{edge}}=i \mid \mathbf{x}_{\pi}) = \frac{\exp(z_{i})}{\sum_{j \in E_{\text{valid}}} \exp(z_{j})},
\end{equation}
where \(E_{\text{valid}} \subseteq E_t \) denotes the set of feasible edges at step \( t \), and \( \pi_{\theta}^{\text{edge}} \) is the learned discrete policy over edge candidates.

Once an edge is selected, the corresponding FAR assignment is modeled as a continuous action sampled from a Beta distribution, which naturally restricts the value to the interval \( (0, 1) \). To parameterize the distribution, the edge logits \( \mathbf{z} \) are concatenated with the policy input \( \mathbf{x}_{\pi} \) and passed through a parameterization module \( f_{\text{far}} \), composed of fully connected layers, yielding the shape parameters \( \alpha \) and \( \beta \):
\begin{equation}
    \alpha, \beta = f_{\text{far}}(\mathbf{x}_{\pi} \Vert \mathbf{z}),
\end{equation}
\begin{equation}
    \pi_{\theta}^{\text{far}}(a^{\text{far}} \mid \alpha, \beta) = \frac{1}{B(\alpha, \beta)} (a^{\text{far}})^{\alpha - 1} (1 - a^{\text{far}})^{\beta - 1}, \quad a^{\text{far}} \in (0, 1),
\end{equation}
where \( \Vert \) denotes the concatenation option, \( B(\alpha, \beta) \) is the beta function for normalization, and \( \pi_{\theta}^{\text{far}} \) is the learned continuous policy over FAR values. This probabilistic formulation allows the policy to adaptively determine intensity levels under varying spatial and contextual constraints. To ensure compliance with planning regulations, each land use type is assigned a predefined maximum allowable FAR. The sampled value \( a_t^{\text{far}} \in (0, 1) \) is then rescaled by multiplying it with the corresponding upper bound, yielding a valid FAR that respects regulatory limits for the target land use type.

To train this compound policy, stochastic action sampling is adopted to promote exploration during learning: the edge is sampled from the masked categorical distribution \( \pi_{\theta}^{\text{edge}} \), and the FAR is drawn from the Beta distribution \( \pi_{\theta}^{\text{far}} \). These actions, along with observed rewards and value estimates, are used to compute policy advantages and update the networks via PPO, as described in Section~\ref{preliminary:DRL}. At inference time, deterministic strategies are applied to ensure stable and high-quality decisions. The edge with the highest probability is selected, and the FAR is assigned as the expected value of the learned Beta distribution. This separation—stochastic exploration during training and deterministic exploitation during inference—strikes a balance between robustness training and consistent deployment performance in generating TOD plans.

\subsection{Value Network}
\label{sec:value_network}

Complementary to the policy network, the value network estimates the expected return of the current planning context to support policy optimization. We first summarize features of the spatial graph by obtaining mean aggregated embeddings of nodes and edges. To obtain fixed-size representations from variable-size graphs, we apply a masked mean aggregation. For edge features, this is computed as:
\begin{equation}
    \mathbf{e}_{\text{mean}} = \frac{\sum_{i=1}^{N_{\text{edge}}} \mathbf{e}_i \cdot \text{mask}_i}{\sum_{i=1}^{N_{\text{edge}}} \text{mask}_i},
\end{equation}
where \( \text{mask}_i \in \{0,1\} \) indicates the validity of edge \( i \). A similar aggregation is used to compute \( \mathbf{v}_{\text{mean}} \), the mean node representation. These operations ensure robustness to varying spatial structures and filter out invalid elements.

The input to the value network, denoted as \( \mathbf{x}_{v} \), is constructed by concatenating the mean node and edge embeddings \( \mathbf{v}_{\text{mean}} \), \( \mathbf{e}_{\text{mean}} \), planning progress vector \( \mathbf{s}_{\text{progress}} \), and the target-aware embedding \( \mathbf{v}_{\text{target}} \). This input captures both macro-level development indicators and the spatial-functional context of the planning state.

The resulting value input \( \mathbf{x}_{v} \) is passed through a value estimator \( f_{\text{value}} \), which consists of fully connected layers, to generate the scalar value prediction:
\begin{equation}
    V_{\theta}(\mathbf{x}_{v}) = f_{\text{value}}(\mathbf{x}_{v}),
\end{equation}
where \( V_{\theta} \) represents the estimated cumulative return. This value is used to compute the advantage function for policy optimization and serves as a baseline to reduce variance during gradient updates in PPO.

The design of the policy and value networks reflects their distinct roles in reinforcement learning. The policy input \( \mathbf{x}_{\pi} \) emphasizes localized spatial interactions for fine-grained decision-making, while the value input \( \mathbf{x}_{v} \) incorporates global contextual features for strategic evaluation. Together, these components form a dual-network structure that enables the PPO agent to generate development strategies that are both context-aware and goal-aligned.

\subsection{Reward Function}\label{method:reward}

To evaluate how well land use patterns align with TOD principles, we design a reward function consisting of two core components: TOD design objectives and development feasibility. These components are assessed within an 800-meter service area around each metro station, providing a balanced framework that integrates long-term planning goals with short-term implementation constraints.

The first component focuses on TOD design objectives, which promote urban form and functionality conducive to transit-oriented planning. We adopt three widely used metrics to quantify TOD performance, as outlined below:

\begin{itemize}
    \item \textbf{Land Use Density (LUDE)}: Density measures the combined intensity of residential, commercial, and office developments near metro stations, reflecting the TOD goals of concentrating population and employment opportunities to enhance transit accessibility, stimulate economic activity, and support sustainable commuting \citep{cervero1997travel, papa2008rail}. We quantify this using a distance-decay weighted metric:
    \begin{equation}
        \mathrm{Density} = \sum_{i=1}^{N_k} \alpha_i \cdot \mathrm{Area}_i \cdot \mathrm{FAR}_i, \quad \alpha_i = \exp\left[-\left(\frac{d_i}{\sigma_d}\right)^2\right],
    \end{equation}
    where \( N_k \) is the number of land parcels of type \( k \), \( d_i \) is the line distance from parcel \( i \) to the nearest metro station, \( \sigma_d \) is the spatial decay parameter, \( \mathrm{Area}_i \) is the area of parcel \( i \) and \( \mathrm{FAR}_i \) is the floor area ratio.

    \item \textbf{Land Use Diversity (LUDI):} A well-mixed land use pattern stabilizes demand across peak and off-peak hours throughout the day and promotes consistent utilization of transit resources \citep{ashik2022investigating, widita2025tod}. We use an entropy-based index to measure diversity:
    \begin{equation}
        \mathrm{Diversity} = \frac{-\sum_{k=1}^{K} P_k \ln P_k}{\ln K},
    \end{equation}
    where \( K \) is the number of land use categories and \( P_k \) is the proportion of category \( k \) within the catchment area of the metro station.

    \item \textbf{Transit Capacity Utilization (TCU):} A core objective of TOD is to increase public transit usage, thereby reducing car dependency \citep{singh2017measuring, su2022deciphering}. To guide land use configurations that maximize transit demand, we develop a simulation model that estimates metro ridership based on the surrounding urban layout. Specifically, we train an XGBoost model on transit usage data from Shanghai \citep{ding2024graph}, using built environment features as input, and apply it to simulate ridership outcomes under different land use scenarios. More details regarding the simulation model are reported in Appendix~\ref{appendix:transit}.
\end{itemize}

The second component of the reward function captures development feasibility (DF), an essential consideration for practical implementation, particularly regarding demolition and reconstruction costs. This is especially important in built-up urban areas, where TOD planning often requires reconfiguring existing urban fabrics rather than developing vacant land. We quantify DF using two indices:

\begin{itemize}
    \item \textbf{Demolition Cost:} We estimate demolition cost based on the total built-up volume to be removed. Specifically, demolition volume is calculated as the sum of: (1) The volume (\(\mathrm{Area} \times \mathrm{FAR}\)) of parcels whose land-use type changes; and (2) The reduced volume (\(\mathrm{Area} \times \Delta\mathrm{FAR}\)) for parcels whose type remains unchanged but whose FAR decreases, divided by the total existing built volume:
    \begin{equation}
        \mathrm{Demolition} =
        \frac{
            \sum_{\substack{\text{type change}}} \mathrm{Area}_i \cdot \mathrm{FAR}_i^{\mathrm{exist}}
            + \sum_{\substack{\text{type unchanged}\\ \Delta\mathrm{FAR}_i<0}} \mathrm{Area}_i \cdot |\Delta\mathrm{FAR}_i|
        }{
            \sum_{j} \mathrm{Area}_j \cdot \mathrm{FAR}_j^{\mathrm{exist}}
        },
    \end{equation}
    where \( \mathrm{FAR}_i^{\mathrm{exist}} \) denotes the existing FAR and \( \Delta\mathrm{FAR}_i = \mathrm{FAR}_i^{\mathrm{new}} - \mathrm{FAR}_i^{\mathrm{exist}} \).

    \item \textbf{Reconstruction Cost:} We estimate reconstruction cost based on the total built-up volume to be added, including two components: (1) The new volume (\(\mathrm{Area} \times \mathrm{FAR}\)) of parcels whose land-use type changes; and (2) The increased volume (\(\mathrm{Area} \times \Delta\mathrm{FAR}\)) for parcels whose type remains unchanged but whose FAR increases, divided by the total regulatory maximum built volume:
    \begin{equation}
        \mathrm{Reconstruction} =
        \frac{
            \sum_{\substack{\text{type change}}} \mathrm{Area}_i \cdot \mathrm{FAR}_i^{\mathrm{new}}
            + \sum_{\substack{\text{type unchanged}\\ \Delta\mathrm{FAR}_i>0}} \mathrm{Area}_i \cdot \Delta\mathrm{FAR}_i
        }{
            \sum_{j} \mathrm{Area}_j \cdot \mathrm{FAR}_j^{\mathrm{max}}
        },
    \end{equation}
    where \( \mathrm{FAR}_j^{\mathrm{max}} \) is the regulatory maximum FAR of parcel \( j \).

\end{itemize}

The final reward is calculated as a weighted sum of the normalized TOD design objectives and development-feasibility components described above. The weights can be determined by domain experts to represent the relative importance assigned to each objective and provide a flexible way to reflect local planning priorities and contextual requirements \citep{singh2017measuring, li2019transit, ibraeva2020transit}. For example, a planning scenario focused on development intensification may assign greater importance to land-use density, whereas another may place greater emphasis on transit capacity utilization, land-use diversity, or development feasibility. The baseline and alternative weight configurations used in this study are described in Section~\ref{subsec:flexibility_analysis}. The specific baseline weight values used for the two Shanghai case studies are reported in Section~\ref{subsec:experimental_setup}; alternative weight configurations exploring trade-offs among the objectives are described in Section~\ref{subsec:flexibility_analysis}, and an additional expert-elicited weight configuration is evaluated in Section~\ref{subsec:workflow_case_demo}.

\section{Experiment Design}
\label{sec:experiment_design}

This section specifies how the DRL4TOD workflow and methodological components introduced in Section~\ref{sec:methodology} were instantiated for the Xintiandi and Jinjing Road station areas in Shanghai. Section~\ref{subsec:case_study} describes the local planning contexts and spatial data, while Section~\ref{subsec:experimental_setup} specifies the land-use categories, FAR constraints, objective weights, training settings, and benchmark methods used in the experiments.

As described in Section~\ref{sec:planning_workflow}, the framework is designed to be configured for different station areas using locally harmonized spatial data, planning requirements, objective weights, and, where applicable, locally recalibrated ridership information; the Shanghai cases provide a concrete application of this general workflow. Specifically, the parcel--road--station graph introduced in Section~\ref{sec:graph_representation} was constructed from Shanghai metro-station locations, parcel geometries, road networks, and existing land-use and FAR data, representing the spatial structure of the two contrasting station-area contexts (Section~\ref{subsec:case_study}). The land-use and FAR actions defined by the MDP in Section~\ref{method:mdp} correspond to nine land-use categories, each constrained by a maximum FAR drawn from Shanghai's planning guidelines, and the multi-objective reward formulated in Section~\ref{method:reward} was assigned weights selected from the literature and domain-expert advice to represent the relative importance of the planning objectives, with the transit-capacity-utilization component calibrated using a ridership model trained on Shanghai data (Section~\ref{subsec:experimental_setup}). The resulting policy and value networks (Sections~\ref{sec:policy_function} and~\ref{sec:value_network}) were trained via PPO and benchmarked against a set of rule-based and optimization-based baselines operating under the same land-use and FAR constraints; the optimization-based methods further adopt the same reward function as DRL4TOD as their optimization objective, and all methods are evaluated using the same set of performance metrics, as detailed in the remainder of this section.

\subsection{Case Study}
\label{subsec:case_study}

In recent years, Shanghai has developed one of the largest urban rail transit networks in the world. However, land development around metro stations has not always been well coordinated with transit infrastructure, leading to limited functional agglomeration near transit nodes and lower-than-expected transit utilization \citep{liu2024station}. In response, Shanghai’s planning policies have increasingly emphasized development intensification around metro stations, making it a representative setting for studying TOD planning. Meanwhile, many station areas in Shanghai are already heavily built up, leaving limited vacant land for new development. In these contexts, TOD planning often requires reconfiguring existing land-use patterns rather than allocating new development only on vacant parcels. This makes development feasibility, particularly reconstruction cost, an important constraint in TOD optimization.

\begin{figure}[!htbp]
  \centering
  \includegraphics[width=\textwidth]{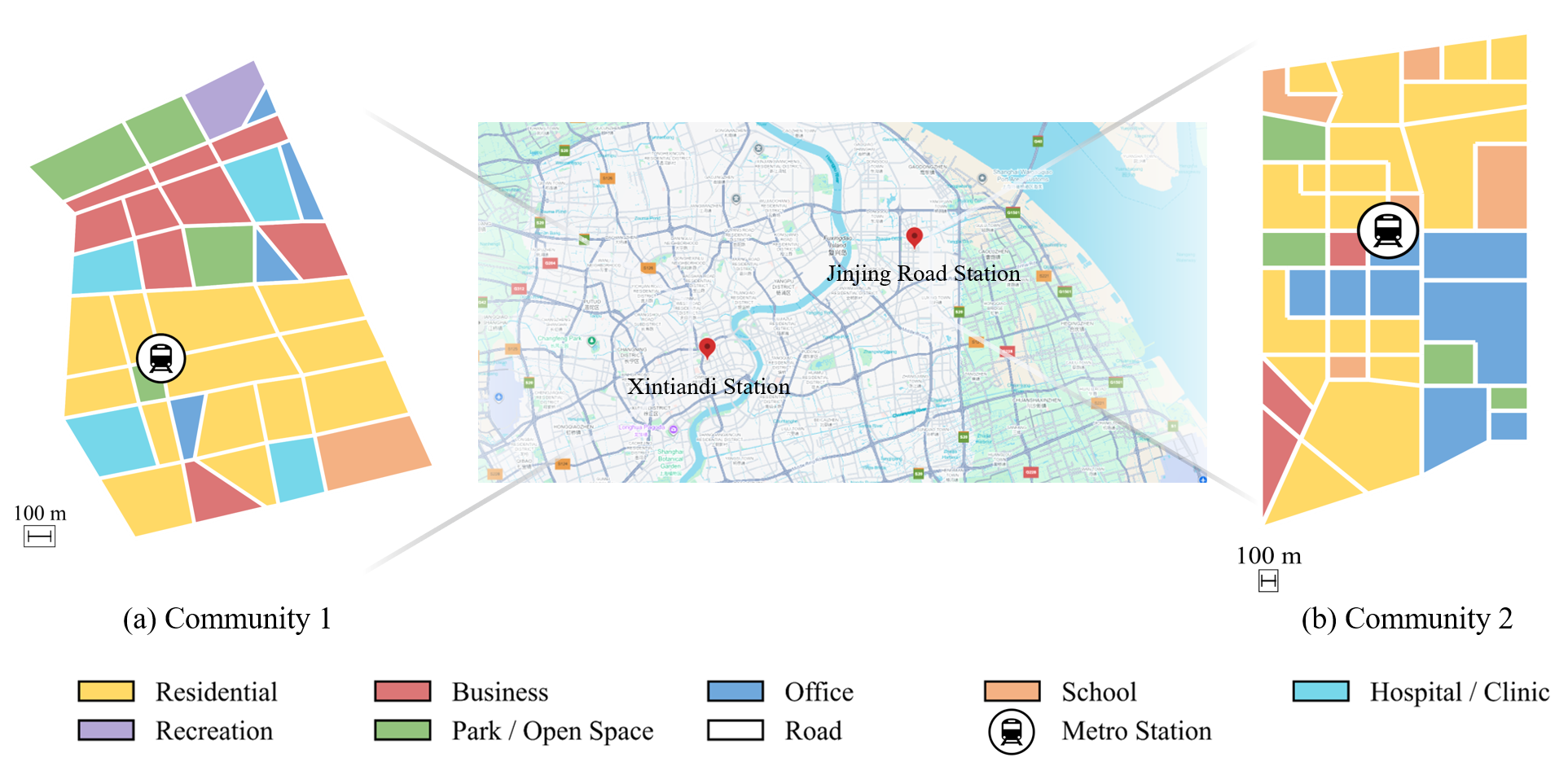}
  \vspace{-10pt}
  \caption{Spatial layouts and functional zoning of the two selected case study communities: (a) Community~01—Xintiandi Station (central urban context) and (b) Community~02—Jinjing Road Station (suburban context). Colors indicate functional land-use categories.}
  \label{fig:case_study_layouts}
\end{figure}

To evaluate the effectiveness and generalizability of the proposed method, two representative communities surrounding metro stations in Shanghai were selected as case studies: Xintiandi Station (Community~01) and Jinjing Road Station (Community~02). Their spatial layouts and functional zoning are shown in Figure~\ref{fig:case_study_layouts}. Community~01, centered on Xintiandi Station, represents a central, highly mixed-use context, whereas Community~02, centered on Jinjing Road Station, represents a more peripheral area dominated by residential and office uses. These contrasting settings allow the framework to be evaluated across different urban contexts. Community boundaries were delineated along major roads, which provide natural physical and administrative divisions between adjacent station areas. Within each boundary, the planning (action) space is deliberately broader than the 800-meter TOD catchment used for evaluation (Section~\ref{method:reward}): rather than restricting reassignable parcels to this buffer, the action space spans the full community. This requires the model to decide not only land-use type and FAR but also which parcels merit inclusion within the service area versus allocation beyond it -- a substantive trade-off, since concentrating a given land-use composition entirely within 800 meters is not always feasible or optimal given parcel geometry and competing objectives, and positioning some uses just outside the buffer can free higher-value parcels near the station for those that benefit most from transit proximity. Restricting the action space to the buffer itself would eliminate this decision and simplify the optimization problem accordingly. The 800-meter radius \citep{cervero2007transit} therefore continues to define where TOD performance is evaluated, while the broader, road-delineated boundary defines the set of parcels from which that evaluation area is populated.

\subsection{Experimental Setup}
\label{subsec:experimental_setup}

This study considers nine typical land-use types commonly found in urban communities: Hospital, School, Clinic, Recreation, Residential, Park, Office, Business, and Open Space. Each type is assigned a maximum allowable floor area ratio according to Shanghai’s urban planning guidelines \citep{ShanghaiGov2020}. These limits, listed in Table~\ref{tab:max_far_by_landuse}, serve as upper bounds in the optimization process. For example, a FAR limit of 1.2 for schools means that the assigned FAR must lie within the range \((0, 1.2)\).

\begin{table}[!htbp]
  \centering
  \footnotesize
  \caption{Maximum allowable floor area ratio by land-use type}
  \begin{tabular}{l|c}
    \toprule
    Land-Use Type & Maximum FAR \\ \midrule
    Hospital \& Clinic & 1.5 \\
    School & 1.2 \\
    Recreation & 2.0 \\
    Residential & 2.5 \\
    Park \& Open Space & 0.0 \\
    Office & 4.0 \\
    Business & 4.0 \\
    \bottomrule
  \end{tabular}
  \label{tab:max_far_by_landuse}
\end{table}

The proposed model jointly optimizes land-use allocation and FAR values to capture the interdependence between spatial arrangement and development intensity in urban planning. The model is trained for 1,000 epochs, with approximately 500 episodes per epoch, generating about 25,000 training samples per episode. The optimization objective is a weighted sum of four normalized reward components introduced in Section~\ref{method:reward}. The weights are selected based on the literature as well as the suggestion of domain experts to balance different planning priorities \citep{singh2017measuring}: LUDE receives the highest weight (0.5) to emphasize efficient use of space in high-demand urban contexts; TCU is weighted at 0.25 to promote transit usage; DF is assigned 0.2 to control for construction cost; and LUDI is given 0.05 to encourage functional diversity without overwhelming other objectives. The weights within LUDE are set to prioritize employment-focused TOD: office density 0.50, business density 0.335, and residential density 0.165, to encourage economic vitality of the community.

To benchmark the performance of the proposed model, we compare it against a set of baseline methods that span distinct optimization paradigms commonly used in TOD and urban land-use planning. Specifically, two rule-based policies provide deterministic planning references, while four optimization-based approaches—Simulated Annealing, Particle Swarm Optimization, Genetic Algorithm, and Adaptive Genetic Algorithm—represent single-solution stochastic search, swarm intelligence, and evolutionary optimization frameworks that are widely adopted in spatial optimization problems \citep{ma2018sustainable, ding2022integrating, liu2020conceptual}. To ensure a fair comparison, all optimization-based methods adopt the same reward function as DRL4TOD.

\begin{itemize}
    \item \textbf{Rule-Based Policies.} The \textit{Rule-Based Centralized Policy (RBCP)} prioritizes development around the community center by allocating land uses to parcels with higher centrality scores, defined based on normalized distance to the center. Floor area ratio values are assigned proportionally to this centrality measure, reflecting conventional planning principles that favor higher density in central locations. In contrast, the \textit{Rule-Based Decentralized Policy (RBDP)} promotes spatial dispersion by sequentially selecting parcels that maximize distance from previously assigned sites. FAR values are determined based on the degree of spatial separation, encouraging more distributed development patterns. Together, these policies represent interpretable yet non-optimized planning heuristics.
    \item \textbf{Simulated Annealing (SA).} SA is a single-solution metaheuristic that explores the search space via stochastic perturbations and a temperature-controlled acceptance criterion \citep{kirkpatrick1983optimization}. By probabilistically accepting inferior solutions during early stages, SA mitigates entrapment in local optima, while gradually focusing on exploitation as the temperature decreases. As a trajectory-based search method, SA provides a representative baseline for stochastic optimization without population-based exploration.
    \item \textbf{Particle Swarm Optimization (PSO).} PSO is a swarm intelligence algorithm in which a population of candidate solutions iteratively updates their positions based on both individual experience and collective knowledge \citep{kennedy1995particle}. Each particle represents a complete land-use and FAR configuration, and the reward function serves as the fitness criterion. PSO has been applied in TOD-related spatial optimization contexts \citep{liu2020conceptual} and provides a population-based baseline with a search mechanism distinct from evolutionary algorithms.
    \item \textbf{Genetic Algorithm (GA).} GA serves as the primary evolutionary optimization baseline. Candidate solutions are encoded as chromosomes representing land-use assignments and FAR values, with the reward function acting as the fitness measure. The algorithm evolves the population through selection, crossover, and mutation, balancing exploration and exploitation across generations. As one of the most widely adopted methods in TOD planning research \citep{ma2018sustainable, ding2022integrating}, GA provides a strong and well-established benchmark.
    \item \textbf{Adaptive Genetic Algorithm (AGA).} AGA extends the standard GA by dynamically adjusting crossover and mutation rates according to population diversity \citep{srinivas2002adaptive}. Specifically, mutation rates are increased when diversity decreases to prevent premature convergence, and reduced when diversity is sufficient to refine promising solutions. This adaptive mechanism enhances search robustness in high-dimensional optimization problems. We include AGA as a representative of enhanced evolutionary approaches that incorporate self-adaptive strategies, complementing the fixed-parameter GA and reflecting recent developments in TOD optimization research \citep{kong2025underground}.
\end{itemize}

This combination of baselines enables a systematic comparison across deterministic heuristics, single-solution stochastic search, and population-based optimization, thereby providing a comprehensive evaluation of DRL4TOD against established planning and optimization approaches.

Each stochastic experiment is repeated five times, and the reported results are averaged across these runs. One-sided t-tests are used to assess whether DRL4TOD achieves higher performance than the comparison methods, with $^{*}p<0.1$ and $^{**}p<0.05$.

\section{Results}
\label{sec:results}

This section presents the empirical evaluation of the proposed DRL4TOD framework. The results are structured into four analyses. Section~\ref{subsec:performance_comparison} benchmarks DRL4TOD against rule-based and optimization-based approaches, evaluating its performance in generating land-use and FAR configurations under the same planning objectives and constraints. Section~\ref{subsec:ablation_analysis} examines the contribution of the main model components, with particular attention to station information and graph-based spatial representation. Section~\ref{subsec:flexibility_analysis} evaluates how changes in planning priorities affect the generated configurations and the trade-offs among density, diversity, transit capacity utilization, and development feasibility. Finally, Section~\ref{subsec:workflow_case_demo} presents an expert-informed application of the planner-guided workflow, in which a panel with backgrounds in transportation and land-use planning specifies station-specific objective weights and reviews a generated plan against considerations beyond the quantitative objectives. A complementary preservation-constrained analysis is reported in Appendix~\ref{appendix:preservation_scenario} to demonstrate how DRL4TOD can support station-area redevelopment when selected parcels must be preserved by fixing those parcels and optimizing the remaining area.

\subsection{Performance Comparison}
\label{subsec:performance_comparison}

\begin{figure}[!htb]
  \centering
  \includegraphics[width=\textwidth]{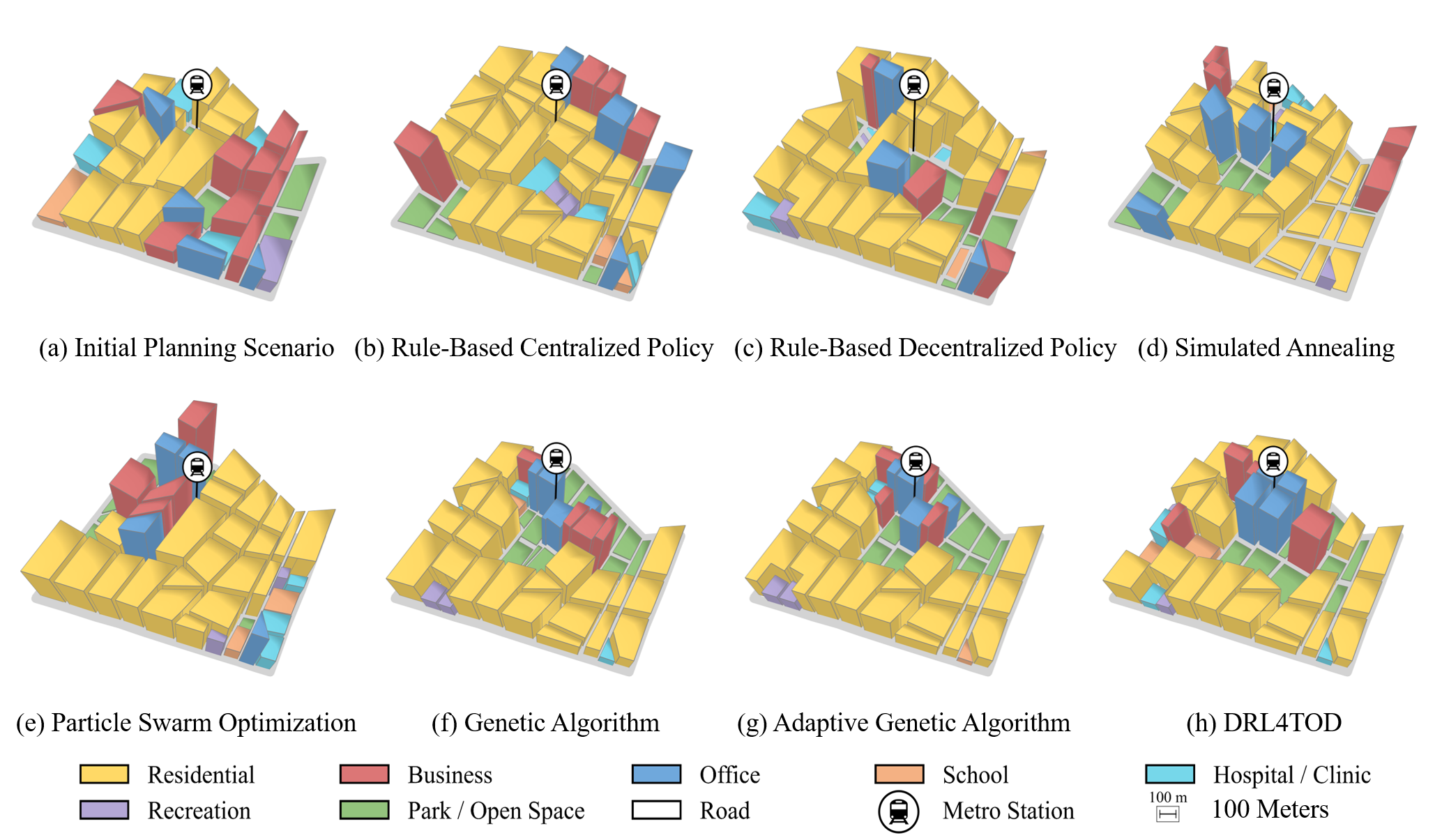}
  \vspace{-8pt}
  \caption{3D FAR distribution under different planning methods at Xintiandi Station. Taller blocks represent higher FAR values.}
  \label{fig:xintiandi_3d_far}
\end{figure}

\begin{figure}[!htb]
  \centering
  \includegraphics[width=\textwidth]{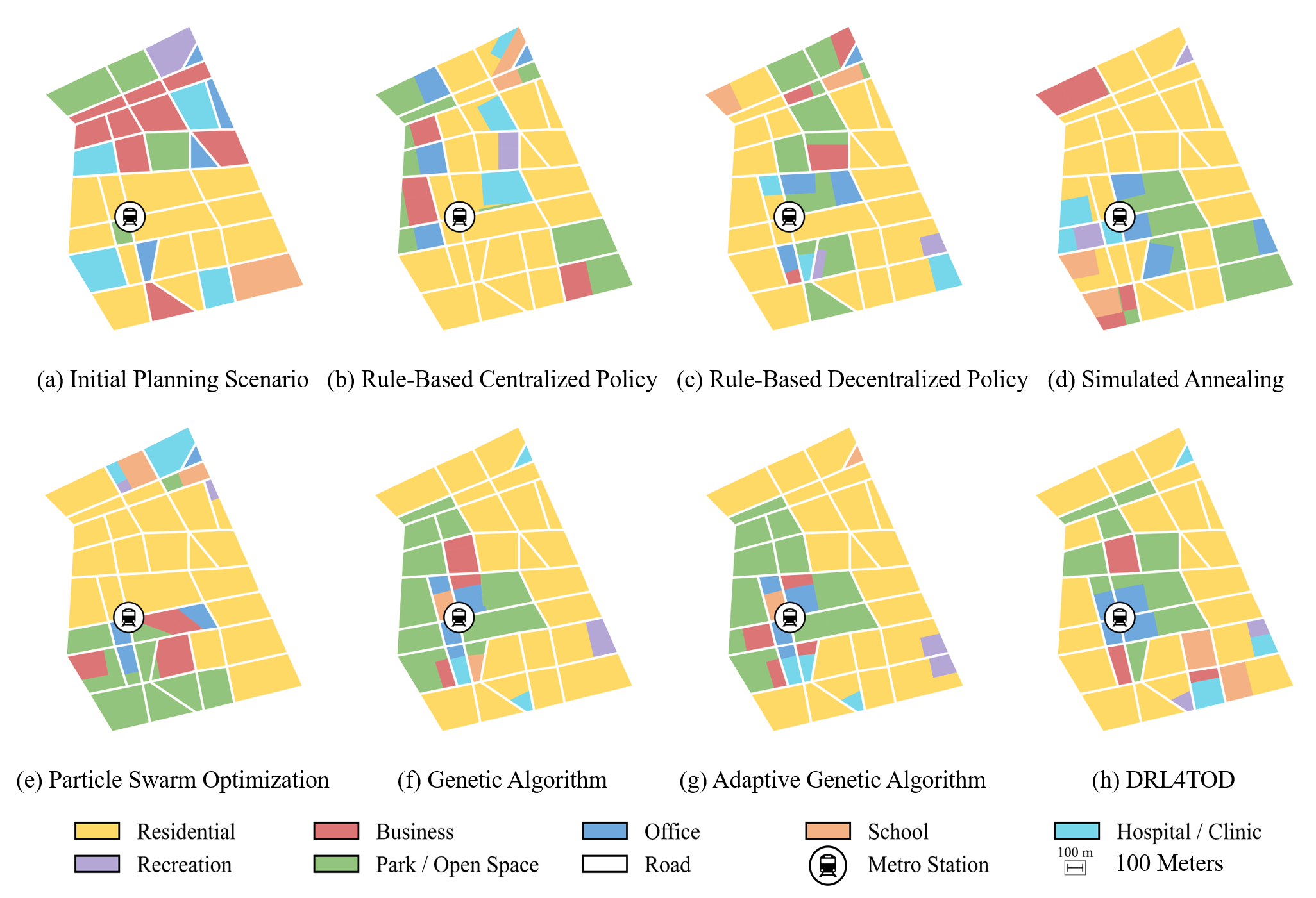}
  \vspace{-8pt}
  \caption{2D land use allocation patterns for Xintiandi Station under different planning methods.}
  \label{fig:xintiandi_2d_allocation}
\end{figure}

The proposed DRL4TOD framework is rigorously compared against seven benchmark planning approaches: the existing Initial Plan, two rule-based heuristics, and four optimization-based methods--Simulated Annealing, Particle Swarm Optimization, Genetic Algorithm, and Adaptive Genetic Algorithm. Table~\ref{tab:tod_performance} summarizes five key TOD-relevant metrics for two representative metro station catchments: Total Reward, Land Use Density, Land Use Diversity, Transit Capacity Utilization, and Development Feasibility. DRL4TOD substantially outperforms the baseline models in Total Reward for both cases, highlighting its robustness in generating TOD-aligned plans across diverse urban contexts. The framework effectively produces spatially concentrated, diverse, and transit-friendly communities with high development feasibility. For an intuitive view of the results, Figures~\ref{fig:xintiandi_3d_far} and~\ref{fig:xintiandi_2d_allocation} depict the 3D FAR distributions and 2D land-use allocations across all methods at Xintiandi Station.

\begin{table}[!htbp]
  \centering \footnotesize
  {
  \caption{Quantitative performance comparison of DRL4TOD and baseline methods across two metro stations. Best results for each station are bold-underlined, and $^{*}$/$^{**}$ denote statistical significance at $p<0.1$ and $p<0.05$, respectively.}
  \begin{tabular}{c|c|ccccc}
    \toprule
    \multirow{2}{*}{Station} & \multirow{2}{*}{Model} & \multicolumn{5}{c}{Metrics} \\
    & & Total Reward & LUDE & LUDI & TCU & DF \\
    \midrule
    \multirow{8}{*}{Xintiandi}
    & Initial Plan & 0.1757 & 0.1732 & 0.7858 & 0.1993 & - \\
    & RBCP         & 0.2521 & 0.2251 & 0.7018 & 0.1728 & 0.3064 \\
    & RBDP         & 0.1886 & 0.0753 & 0.7071 & 0.1697 & 0.3658 \\
    & SA           & 0.4507 & 0.5751 & 0.8333 & 0.1680 & 0.3975 \\
    & PSO          & 0.4542 & 0.6122 & 0.7027 & 0.2172 & 0.2935 \\
    & GA           & 0.5448 & 0.7418 & 0.8138 & 0.2045 & 0.4103 \\
    & AGA          & 0.5826 & 0.8216 & 0.8103 & 0.2183 & 0.3834 \\
    & DRL4TOD      & \textbf{\underline{1.0371}}$^{**}$ & \textbf{\underline{1.6965}}$^{**}$ & \textbf{\underline{0.8385}}$^{**}$ & \textbf{\underline{0.2272}}$^{**}$ & \textbf{\underline{0.4505}}$^{**}$ \\
    \midrule
    \multirow{8}{*}{Jinjing}
    & Initial Plan & 0.1747 & 0.1464 & 0.7300 & 0.2598 & - \\
    & RBCP         & 0.2230 & 0.1507 & 0.6930 & 0.1595 & 0.3654 \\
    & RBDP         & 0.1880 & 0.1218 & 0.6378 & 0.1756 & 0.2566 \\
    & SA           & 0.2504 & 0.2268 & 0.6521 & 0.1762 & 0.3017 \\
    & PSO          & 0.2799 & 0.2124 & 0.6638 & 0.1849 & 0.4715 \\
    & GA           & 0.3371 & 0.3147 & 0.6522 & 0.1867 & 0.5024 \\
    & AGA          & 0.3473 & 0.3207 & 0.6889 & 0.2036 & 0.5081 \\
    & DRL4TOD      & \textbf{\underline{0.5518}}$^{**}$ & \textbf{\underline{0.6474}}$^{*}$ & \textbf{\underline{0.7409}}$^{**}$ & \textbf{\underline{0.3383}}$^{*}$ & \textbf{\underline{0.5323}}$^{**}$ \\
    \bottomrule
  \end{tabular}
  \label{tab:tod_performance}
}
\end{table}

DRL4TOD demonstrates clear superiority in generating high-density development around the metro station. For the Xintiandi catchment, it achieves a LUDE of 1.6965, approximately twice the score of the strongest optimization baseline (AGA with 0.8216). Similarly, for Jinjing Road, DRL4TOD achieves a LUDE of 0.6474, substantially outperforming the best optimization baseline AGA (0.3207) by a factor of approximately two. As shown in Fig.~\ref{fig:xintiandi_2d_allocation}(h), DRL4TOD strategically concentrates office and business parcels immediately adjacent to the station, consistent with the LUDE weight priority (office $>$ business $>$ residential). Complementary uses — including school, hospital, and recreational facilities — are distributed within the catchment, while residential parcels occupy the periphery. The 3D visualization (Fig.~\ref{fig:xintiandi_3d_far}(h)) confirms this pattern: the tallest, highest-intensity office and business blocks cluster directly around the transit node, with building heights tapering outward toward lower-density residential development. This spatial organization embodies TOD's density principle, ensuring sufficient employment and activity concentration to sustain high-frequency transit service. This spatial pattern is also consistent with the underlying economic context of the station area. As described in Section~\ref{subsec:case_study}, Xintiandi is located near Shanghai's city center within a highly mixed-use urban fabric and serves as a major commercial hub. The LUDE weighting used in this study (Section~\ref{method:reward}) deliberately prioritizes office and business density over other objectives, reflecting TOD principles that treat employment concentration near transit stations as a driver of both ridership and economic vitality. Applied to Xintiandi's centrally located, commercially oriented context, this weighting produces the outcome that would be expected of a well-designed TOD plan in a high-demand district: office and business parcels concentrated at the highest feasible intensity immediately around the station. This indicates that DRL4TOD's output reflects not only reward-function optimization in the abstract, but also a response calibrated to the station area's real-world development context.

DRL4TOD achieves the highest LUDI across both stations, reaching 0.8385 at Xintiandi -- comparable to SA (0.8333) but combined with substantially higher density and transit performance -- and 0.7409 at Jinjing Road, surpassing the initial plan (0.7300) and the best-performing baseline, AGA (0.6889). As shown in Fig.~\ref{fig:xintiandi_2d_allocation}(h), DRL4TOD co-locates office, business, residential, healthcare, and school parcels within walking distance of the station, concentrating complementary activities that support all-day transit demand across different trip purposes and time periods. This diversity is achieved alongside, rather than at the cost of, transit performance: DRL4TOD attains the highest TCU among all methods at both stations, reaching 0.3383 at Jinjing Road (a 30.2\% improvement over the initial plan's 0.2598, and well above AGA's 0.2036) and 0.2272 at Xintiandi (exceeding AGA at 0.2183 and PSO at 0.2172). The 3D visualization (Fig.~\ref{fig:xintiandi_3d_far}(h)) reinforces this finding: the clustering of high-intensity mixed uses around the transit node generates a built-environment configuration that sustains substantially higher simulated passenger demand than any baseline configuration.

In addition to enhancing TOD performance, DRL4TOD achieves the highest DF scores across both stations, demonstrating that superior TOD outcomes are attained without disproportionate disruption to the existing urban fabric. For Xintiandi, it attains a DF of 0.4505, exceeding the best optimization baseline GA (0.4103) by 9.8\%. For Jinjing Road, it achieves a DF of 0.5323, a 4.8\% improvement over AGA (0.5081). These results confirm that DRL4TOD's sequential, context-aware optimization not only maximizes TOD performance but does so through configurations that require comparatively conservative intervention in the existing built stock — a practically important property for urban renewal contexts where demolition and reconstruction costs are real planning constraints.

By contrast, the baseline methods fail to adequately address the multi-objective nature of TOD planning. The initial plan (Figs.~\ref{fig:xintiandi_2d_allocation}(a) and~\ref{fig:xintiandi_3d_far}(a)) concentrates residential parcels directly adjacent to the station while displacing higher-intensity employment and commercial uses to the periphery — an arrangement that provides residential transit access but limits employment clustering and mixed-use integration, conflicting with TOD's density and diversity principles. The rule-based approaches (Figs.~\ref{fig:xintiandi_3d_far}(b,c) and~\ref{fig:xintiandi_2d_allocation}(b,c)) produce scattered land-use patterns that fail to leverage transit accessibility strategically: the centralized policy places large residential blocks immediately around the station with limited functional diversity, while the decentralized policy distributes uses broadly without the density concentration that TOD requires. Among the optimization-based methods, performance improves progressively in the order SA $<$ PSO $<$ GA $<$ AGA, consistent with the incremental benefit of population-based and adaptive evolutionary search over single-solution stochastic methods. SA (Figs.~\ref{fig:xintiandi_3d_far}(d) and~\ref{fig:xintiandi_2d_allocation}(d)) achieves relatively high diversity (LUDI 0.8333 at Xintiandi) but produces a spatially diffuse pattern without clear concentration of high-intensity uses near the transit node, reflecting the limited spatial reasoning capacity of single-solution trajectory search. PSO (Figs.~\ref{fig:xintiandi_3d_far}(e) and~\ref{fig:xintiandi_2d_allocation}(e)) improves on density but sacrifices functional diversity (LUDI 0.7027 at Xintiandi), with large homogeneous parcels dominating the catchment at the cost of mixed-use balance. GA and AGA (Figs.~\ref{fig:xintiandi_3d_far}(f,g) and~\ref{fig:xintiandi_2d_allocation}(f,g)) produce progressively better-balanced configurations — AGA's adaptive mutation mechanism enables more consistent clustering of office and business uses near the station than standard GA — yet both fall considerably short of DRL4TOD's density levels and the spatial coherence visible in Fig.~\ref{fig:xintiandi_3d_far}(h). Overall, DRL4TOD achieves the best balance of TOD's density, diversity, and accessibility principles across all methods and both stations, as confirmed by both the numerical results in Table~\ref{tab:tod_performance} and the spatial patterns visible in Figures~\ref{fig:xintiandi_3d_far} and~\ref{fig:xintiandi_2d_allocation}.

\subsection{Ablation Analysis}
\label{subsec:ablation_analysis}

\begin{figure}[!htb]
  \centering
  \includegraphics[width=\textwidth]{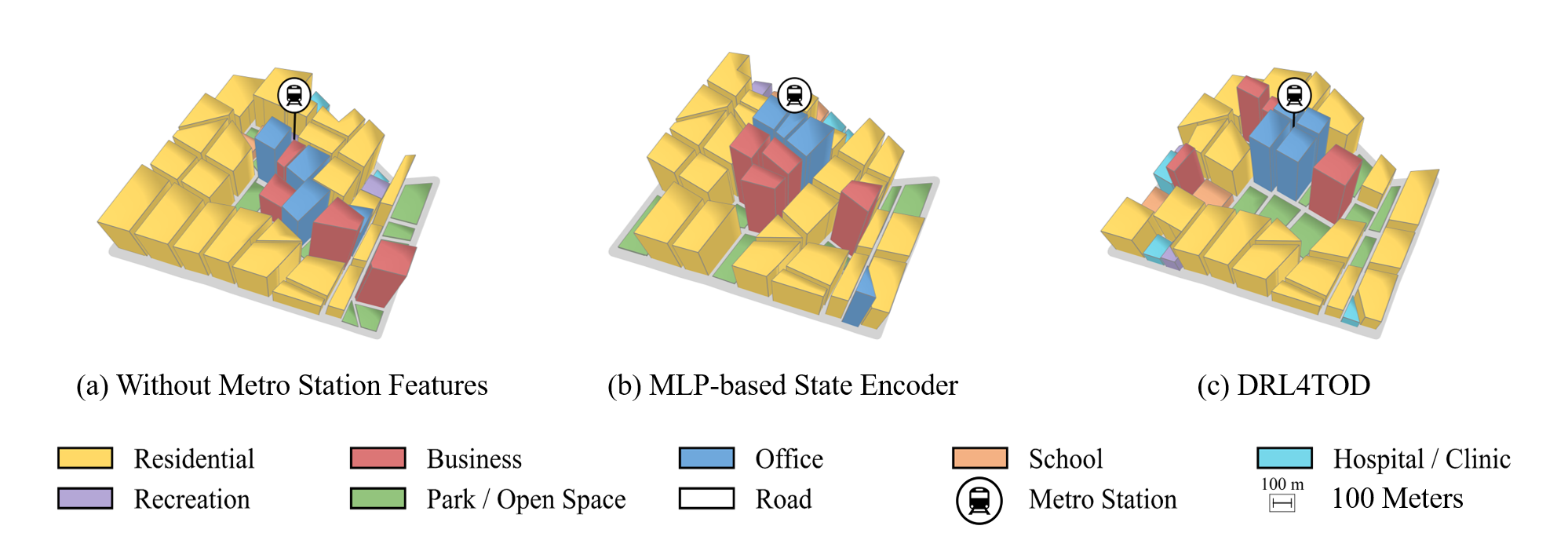}
  \vspace{-20pt}
  \caption{3D visualization of the ablation study results}
  \label{fig:ablation_3d_results}
\end{figure}

\begin{figure}[!htb]
  \centering
  \includegraphics[width=\textwidth]{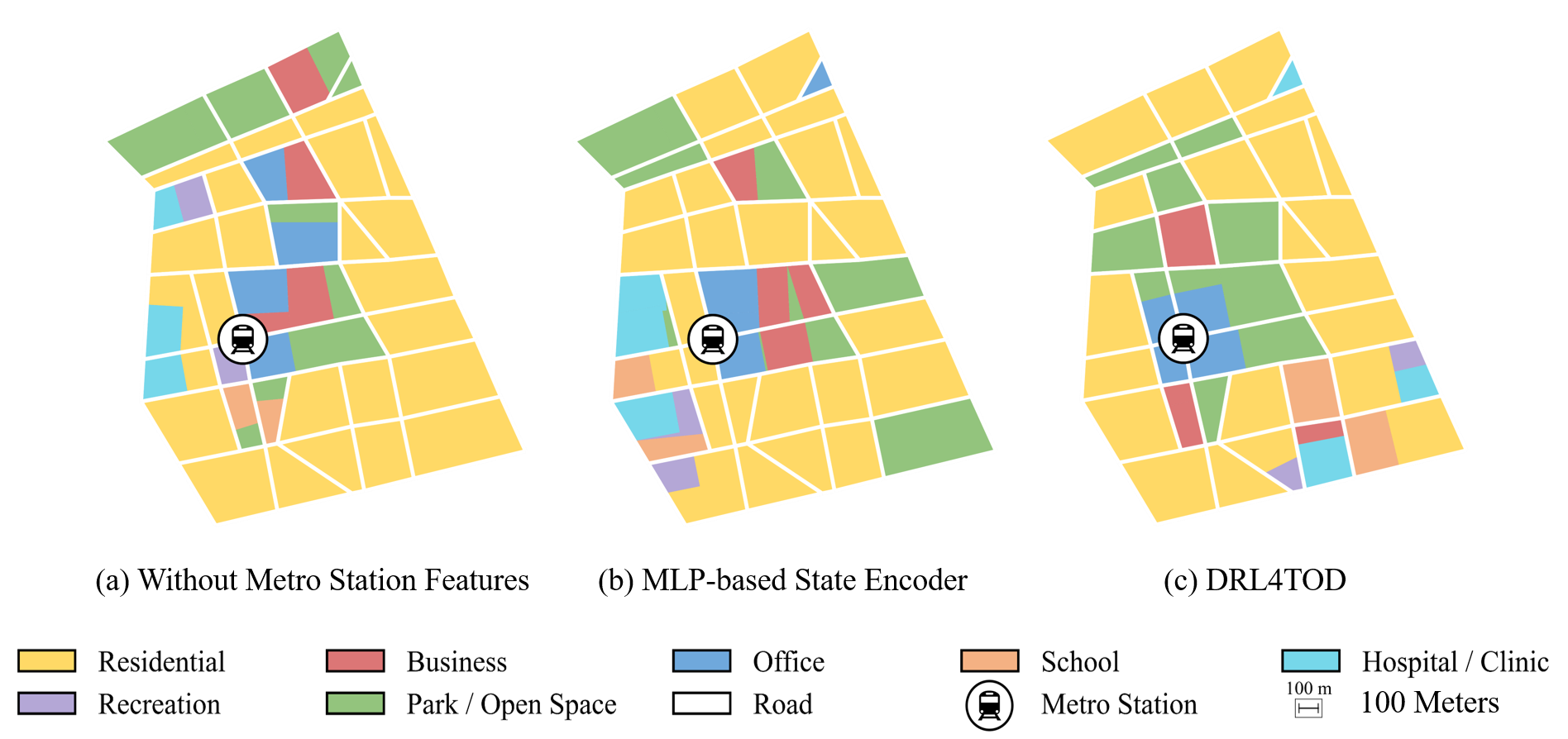}
  \vspace{-20pt}
  \caption{2D visualization of the ablation study results}
  \label{fig:ablation_2d_results}
\end{figure}

To evaluate the contribution of individual model components in DRL4TOD, we conduct an ablation study by systematically modifying or removing critical architectural elements. Table~\ref{table:ablation_performance} reports the quantitative results for the Xintiandi Station case, while Figures~\ref{fig:ablation_3d_results} and~\ref{fig:ablation_2d_results} present the corresponding spatial allocations.

We first examine the impact of incorporating metro station location information into the GNN state encoder. Removing this spatial input leads to a substantial decline in performance, with the total reward decreasing from 1.0371 to 0.5232 and LUDE decreasing by 59.1\% from 1.6965 to 0.6939. As visible in Figures~\ref{fig:ablation_3d_results}(a) and~\ref{fig:ablation_2d_results}(a), the resulting configuration displays a spatially scattered land-use distribution with no coherent concentration of high-intensity functions near the transit node. While office and business parcels are present in the catchment, they are dispersed across the area without the station-proximity ordering that characterizes effective TOD layouts. This dispersed pattern indicates a fundamental failure to leverage transit accessibility as an organizing principle, reaffirming that metro station location features are essential for guiding density clustering and land-use synergy in line with TOD principles.

Next, we assess the importance of spatial feature learning by replacing the GNN-based encoder with a Multilayer Perceptron (MLP). As shown in Table~\ref{table:ablation_performance}, the MLP variant performs worse than the full model across all metrics, achieving a total reward of 0.7984, LUDE of 1.3011, and LUDI of 0.8106. Most notably, it records the lowest DF score among all variants (0.2701), indicating that without graph-based spatial reasoning, the model generates configurations that require substantially more aggressive intervention in the existing built fabric to achieve its density targets. As visible in Figures~\ref{fig:ablation_3d_results}(b) and~\ref{fig:ablation_2d_results}(b), the MLP variant places high-intensity office and business blocks at elevated FAR values but distributes them inconsistently across the catchment, with fragmented spatial logic that does not reflect the topological relationship between individual parcels and the transit node. The 2D layout reveals a mix of uses — including hospital and recreational facilities — appearing near the station without coherent proximity-based organization, suggesting the MLP encodes parcel attributes in isolation rather than in relation to the broader station-area context. These outcomes highlight the superior spatial reasoning capacity of GNNs, which capture the topological interdependencies among parcels that are central to generating realistic and feasible TOD configurations.

In contrast, the DRL4TOD full model, which integrates metro station features with a GNN-based encoder, produces the best performance across all TOD objectives, achieving a total reward of 1.0371, LUDE of 1.6965, LUDI of 0.8385, TCU of 0.2272, and DF of 0.4505. As shown in Figures~\ref{fig:ablation_3d_results}(c) and~\ref{fig:ablation_2d_results}(c), the full model generates a spatially coherent configuration in which the tallest, highest-intensity office and business blocks are concentrated directly around the transit node, with building heights tapering outward toward lower-density residential development. The 2D layout confirms this station-centric organization: office and business parcels occupy the immediate station vicinity, park and open space uses are distributed at the periphery, and complementary facilities including school and hospital are positioned within walking distance of the station. This arrangement simultaneously realizes TOD's density principle through employment concentration near the hub, its diversity principle through the co-location of complementary functions, and its feasibility dimension through a comparatively conservative departure from the existing built stock — as reflected in the highest DF score among all variants. These findings validate the design choices in DRL4TOD and demonstrate that both transit-aware input features and graph-based spatial encoding are necessary conditions for generating high-quality, renewal-oriented TOD plans.

\begin{table}[!htbp]
  \centering
  \footnotesize
  \caption{Ablation study: performance comparison on Xintiandi Station}
  \begin{tabular}{c|ccccc}
    \toprule
    \multirow{2}{*}{Model Variant} & \multicolumn{5}{c}{Performance Metrics} \\
    & Total Reward & LUDE & LUDI & TCU & DF \\
    \midrule
    Without Metro Station Data & 0.5232 & 0.6939 & 0.8365 & 0.2100 & 0.4096 \\
    MLP State Encoder & 0.7984 & 1.3011 & 0.8106 & 0.2132 & 0.2701 \\
    DRL4TOD (Full Model) & \textbf{\underline{1.0371}} & \textbf{\underline{1.6965}} & \textbf{\underline{0.8385}} & \textbf{\underline{0.2272}} & \textbf{\underline{0.4505}} \\
    \bottomrule
  \end{tabular}
  \label{table:ablation_performance}
\end{table}

\subsection{Planning Priority Scenario Analysis}
\label{subsec:flexibility_analysis}

To examine how planning priorities shape DRL4TOD outputs, we construct a set of objective-weight scenarios by varying the relative weights assigned to LUDE, LUDI, TCU, and DF while keeping the spatial data, action space, and training settings unchanged. Each scenario is evaluated using the same performance indicators and spatial diagnostics. Comparing the resulting plans reveals the trade-offs among competing objectives and illustrates how DRL4TOD can generate alternative configurations for density-focused, diversity-focused, transit-oriented, or redevelopment-conscious planning contexts. The outcomes are visualized across four figures: Figures~\ref{fig:weight_3d_lud_ludi_tcu} and \ref{fig:weight_2d_lud_ludi_tcu} present a baseline scenario alongside results from modifications to the LUDE, LUDI, and TCU weights. Figures \ref{fig:development_feasibility_3D} and \ref{fig:development_feasibility_2D} illustrate the initial plan, baseline, and the outcome of increasing the DF weight. Numerical results are reported in two tables: Table~\ref{tab:lud_weights} (LUDE) and Table~\ref{tab:ludi_tcu_weights} (LUDI, TCU, and DF). %

\subsubsection{Impact of Land-Use Density Weights}
\label{subsec:impact_lud}

Land-use density plays a central role in shaping the intensity and concentration of urban development around metro stations. To assess its effect, we reweighted the density components to place much greater emphasis on residential density, increasing its weight from 0.0825 to 0.4050 while reducing the weights for business and office density. This adjustment significantly influenced both the numerical performance and spatial configuration of the output.

As shown in Table~\ref{tab:lud_weights}, the modified weighting yields a pronounced rise in residential density (from 0.0141 to 0.0452, a \(\approx\)220.6\% increase) and a modest increase in business density (from 0.0726 to 0.0890), while office density drops sharply from 3.3397 to 0.7655. The DF score also decreases from 0.4505 to 0.3711, indicating that a residential-focused configuration requires more substantial intervention in the existing built fabric to realize its density targets. These variations highlight the model's ability to adapt its optimization priorities in direct response to planner-specified weights, though the trade-off between residential concentration and overall TOD performance is clearly reflected in the numerical outcomes.

\begin{figure}[!htb]
    \centering
    \includegraphics[width=\textwidth]{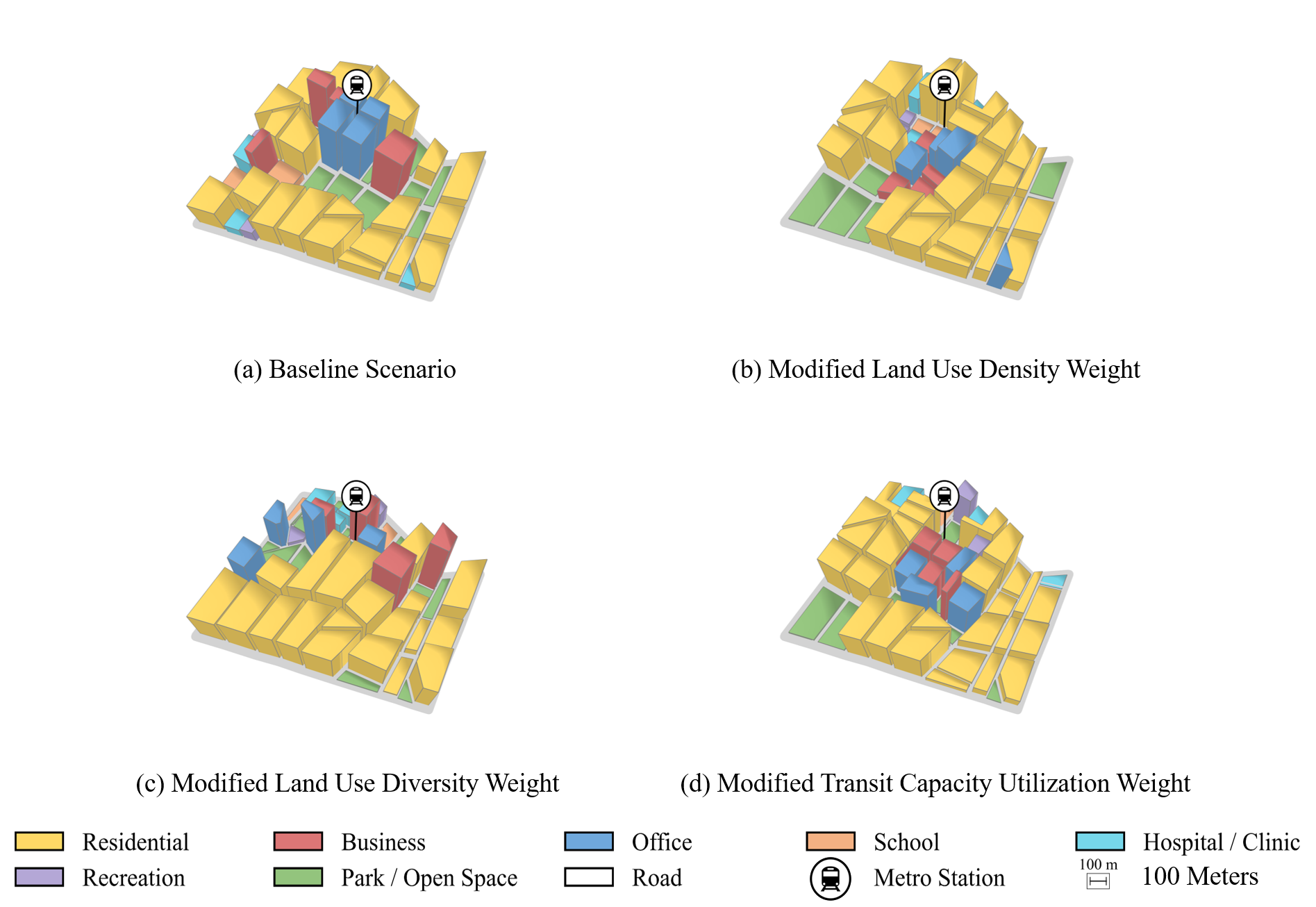}
    \caption{3D spatial planning outcomes under varying TOD objective weights.}
    \label{fig:weight_3d_lud_ludi_tcu}
\end{figure}

\begin{figure}[!htb]
    \centering
    \includegraphics[width=\textwidth]{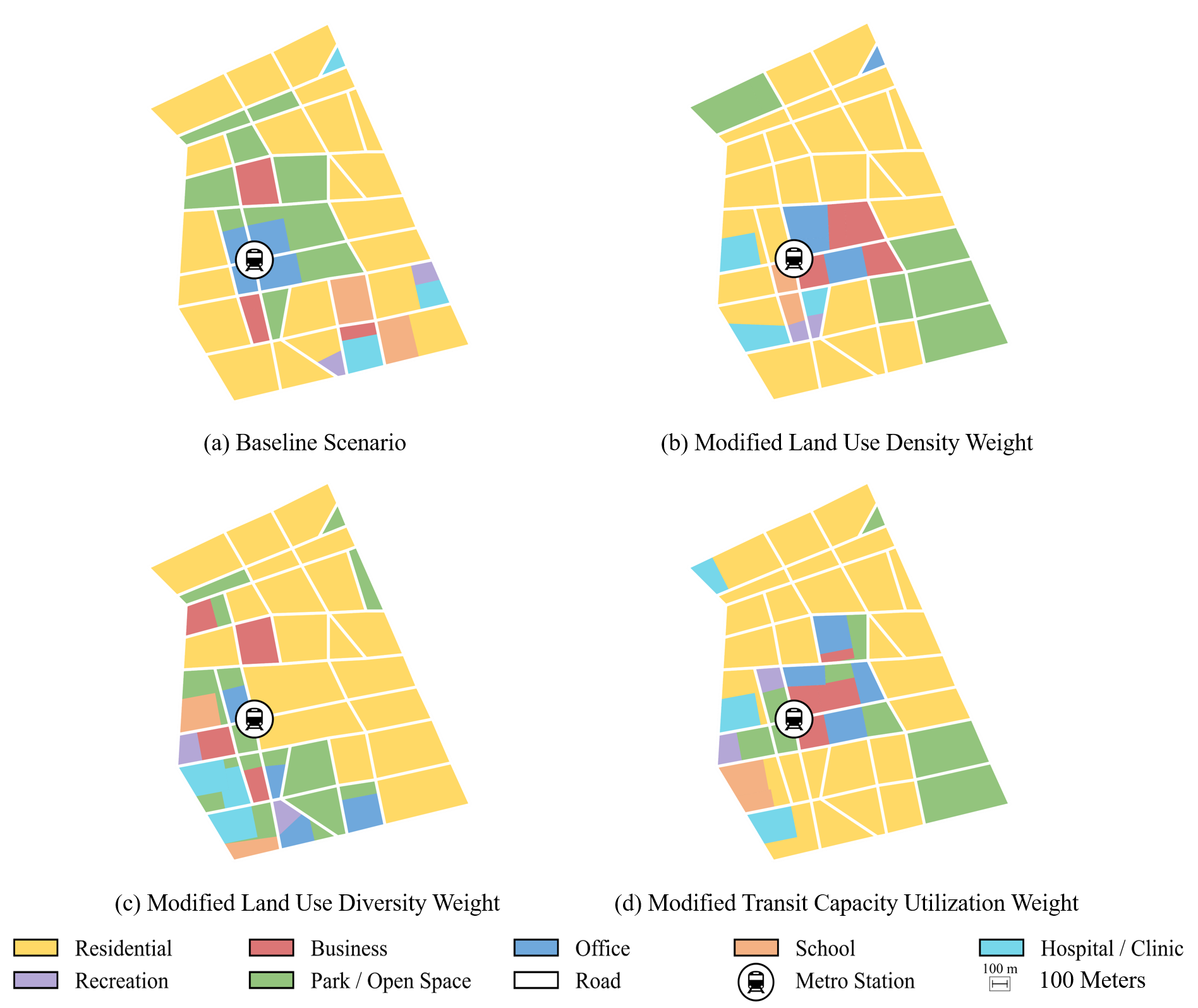}
    \caption{2D spatial planning outcomes under varying TOD objective weights.}
    \label{fig:weight_2d_lud_ludi_tcu}
\end{figure}

The spatial shift is clearly illustrated in Figures~\ref{fig:weight_3d_lud_ludi_tcu}(a,b) and~\ref{fig:weight_2d_lud_ludi_tcu}(a,b). Under the baseline configuration (panel a), tall office and business towers cluster directly around the transit node, with residential and complementary uses occupying the surrounding blocks. Under the modified LUDE weights (panel b), the model reallocates FAR toward residential parcels near the station: the 3D visualization shows a marked reduction in the height and spatial extent of office towers, replaced by a more uniform low-to-mid-rise residential profile across the catchment, while the 2D layout reveals a shift toward larger residential parcels with office and business uses displaced to secondary positions. In practice, these results illustrate how planners can use DRL4TOD to steer development toward resident-focused or employment-focused station areas, with the weight configuration providing a direct and transparent mechanism for expressing planning priorities.

\begin{table}[!htbp]
    \centering \footnotesize
    \caption{Effect of residential density weight on performance metrics. The adjusted weight and its corresponding score are bold-underlined.}
    \begin{tabular}{cccccccc}
        \hline
        & & \multicolumn{3}{c}{LUDE} & \multirow{2}{*}{LUDI} & \multirow{2}{*}{TCU} & \multirow{2}{*}{DF} \\ \cline{3-5}
        & & Residential & Business & Office & & & \\ \hline
        \multirow{2}{*}{Weight} & Baseline & \underline{\textbf{0.0825}} & 0.1675 & 0.25 & 0.05 & 0.25 & 0.2 \\
        & Modified LUDE & \underline{\textbf{0.4050}} & 0.0450 & 0.05 & 0.05 & 0.25 & 0.2 \\\hline
        \multirow{2}{*}{Value} & Baseline & \underline{\textbf{0.0141}} & 0.0726 & 3.3397 & 0.8385 & 0.2272 & 0.4505 \\
        & Modified LUDE & \underline{\textbf{0.0452}} & 0.0890 & 0.7655 & 0.8080 & 0.2653 & 0.3711 \\\hline
    \end{tabular}
    \label{tab:lud_weights}
\end{table}

\subsubsection{Impact of Land Use Diversity Weights}
\label{subsec:impact_ludi}

Land use diversity is a cornerstone of TOD, reflecting the functional balance among different land-use types within the metro station’s service area. In this experiment, we significantly increased the LUDI weight from $0.05$ in the baseline to $0.80$, while reducing the emphasis on the other objectives, to examine how the model prioritizes mixed-use development.

As shown in Table~\ref{tab:ludi_tcu_weights}, this adjustment improved the diversity score from a baseline of 0.8385 to 0.9415 ($+$12.25\%), while LUDE decreased substantially from 1.6965 to 0.3022, reflecting the expected trade-off between functional diversity and employment density concentration. Figures~\ref{fig:weight_3d_lud_ludi_tcu}(c) and~\ref{fig:weight_2d_lud_ludi_tcu}(c) illustrate the corresponding spatial changes. Compared to the baseline scenario in Figures~\ref{fig:weight_3d_lud_ludi_tcu}(a) and~\ref{fig:weight_2d_lud_ludi_tcu}(a), the modified plan introduces a richer and more spatially distributed mix of uses around the metro station: the 3D visualization shows a more fragmented building height profile with multiple use types of varying intensities interspersed across the catchment, rather than the concentrated tower cluster of the baseline, while the 2D layout reveals a finer-grained mosaic of residential, office, business, recreational, park, healthcare, and school parcels distributed around the station. These results highlight that a small baseline weight on LUDI is already sufficient to ensure a balanced land-use mix, while excessively prioritizing diversity reduces density advantages. This finding emphasizes the importance of balancing TOD dimensions rather than maximizing any single principle in isolation.

\begin{table}[htbp!]
    \centering \footnotesize
    \caption{Effect of LUDI, TCU, and DF weights on planning performance. The adjusted weights and corresponding scores are bold-underlined.}
    \begin{tabular}{cccccc}
        \hline
        & & LUDE & LUDI & TCU & DF \\ \hline
        \multirow{4}{*}{Weight} & Baseline & 0.5 & \underline{\textbf{0.05}} & \underline{\textbf{0.25}} & \underline{\textbf{0.2}} \\
        & Modified LUDI & 0.05 & \underline{\textbf{0.8}} & 0.05 & 0.1 \\
        & Modified TCU  & 0.1 & 0.05 & \underline{\textbf{0.8}} & 0.05 \\
        & Modified DF & 0.1 & 0.05 & 0.05 & \underline{\textbf{0.8}} \\ \hline
        \multirow{4}{*}{Value} & Baseline & 1.6965 & \underline{\textbf{0.8385}} & \underline{\textbf{0.2272}} & \underline{\textbf{0.4505}} \\
        & Modified LUDI & 0.3022 & \underline{\textbf{0.9415}} & 0.2013 & 0.4876 \\
        & Modified TCU & 0.4087 & 0.8707 & \underline{\textbf{0.2995}} & 0.4407 \\
        & Modified DF & 0.0584 & 0.7762 & 0.1993 & \underline{\textbf{0.5837}} \\ \hline
    \end{tabular}
    \label{tab:ludi_tcu_weights}
\end{table}

\subsubsection{Impact of Transit Capacity Utilization Weights}
\label{subsec:impact_tcu}

Transit capacity utilization provides a direct measure of how effectively land-use configurations support metro ridership. In this experiment, we significantly increased the TCU reward weight from 0.25 to 0.8, reducing the emphasis on other objectives to prioritize transit efficiency.

As reported in Table~\ref{tab:ludi_tcu_weights}, this adjustment led to a substantial increase in the TCU score from a baseline of 0.2272 to 0.2995 ($+$31.9\%), with LUDE decreasing to 0.4087 and DF remaining stable at 0.4407 — suggesting that transit-optimized configurations achieve higher ridership support without substantially increasing redevelopment disturbance relative to the baseline. The visual evidence in Figures~\ref{fig:weight_3d_lud_ludi_tcu}(d) and~\ref{fig:weight_2d_lud_ludi_tcu}(d) illustrates this shift clearly. Compared to the baseline scenario in Figures~\ref{fig:weight_3d_lud_ludi_tcu}(a) and~\ref{fig:weight_2d_lud_ludi_tcu}(a), the modified TCU layout concentrates office and business blocks in a tight cluster immediately adjacent to the metro station, as visible in the 3D visualization where high-rise commercial and office towers form a clustered transit-proximate grouping, while residential development occupies the outer ring at lower intensities. The 2D layout confirms this pattern, with a more contiguous band of transit-supportive uses — including office, business, and hospital parcels — arranged around the station node. These results demonstrate that DRL4TOD can effectively optimize for transit ridership as a primary objective, generating configurations that concentrate the land-use types most associated with high transit demand in the locations most accessible to the metro system.

\subsubsection{Impact of Development Feasibility Weights}
\label{subsec:impact_df}

Development feasibility represents the constructability and ecological sustainability of proposed land use configurations. Prioritizing this metric encourages a more conservative and realistic development approach, helping planners avoid excessive construction in less suitable areas and preserve land for long-term resilience. This aligns with broader TOD and sustainable urban planning goals, such as minimizing environmental impact, maintaining ecological buffers, and ensuring realistic implementation.

Table~\ref{tab:ludi_tcu_weights} reports the results of significantly increasing the DF weight from 0.20 to 0.80. This adjustment led to a substantial improvement in the DF score from a baseline of 0.4505 to 0.5837 ($+$29.57\%), indicating that the modified plan retains a considerably larger proportion of the existing built stock unchanged. However, this gain comes at the cost of significant trade-offs across the remaining TOD objectives: LUDE decreased sharply from 1.6965 to 0.0584, LUDI dropped from 0.8385 to 0.7762, and TCU declined from 0.2272 to 0.1993. These outcomes reflect the fundamental tension between conserving the existing urban fabric and achieving the density, diversity, and transit performance goals that define effective TOD — a tension that is particularly acute in mature station areas where the existing land-use pattern may be poorly aligned with TOD objectives. The baseline demolition ratio is 0.3389, while the modified DF's demolition ratio is 0.4719.

\begin{figure}[!htbp]
    \centering
    \includegraphics[width=\textwidth]{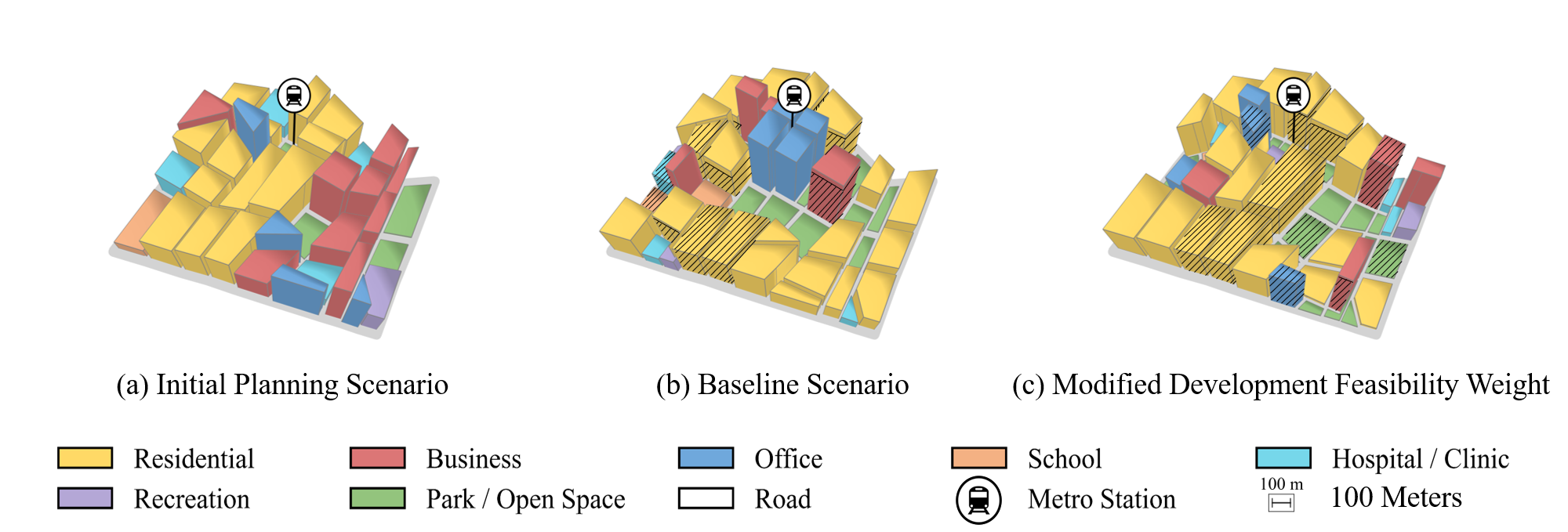}
    \caption{3D spatial planning outcomes under development feasibility weight adjustment. Hatched parcels indicate preserved areas.}
    \label{fig:development_feasibility_3D}
\end{figure}

\begin{figure}[!htbp]
    \centering
    \includegraphics[width=\textwidth]{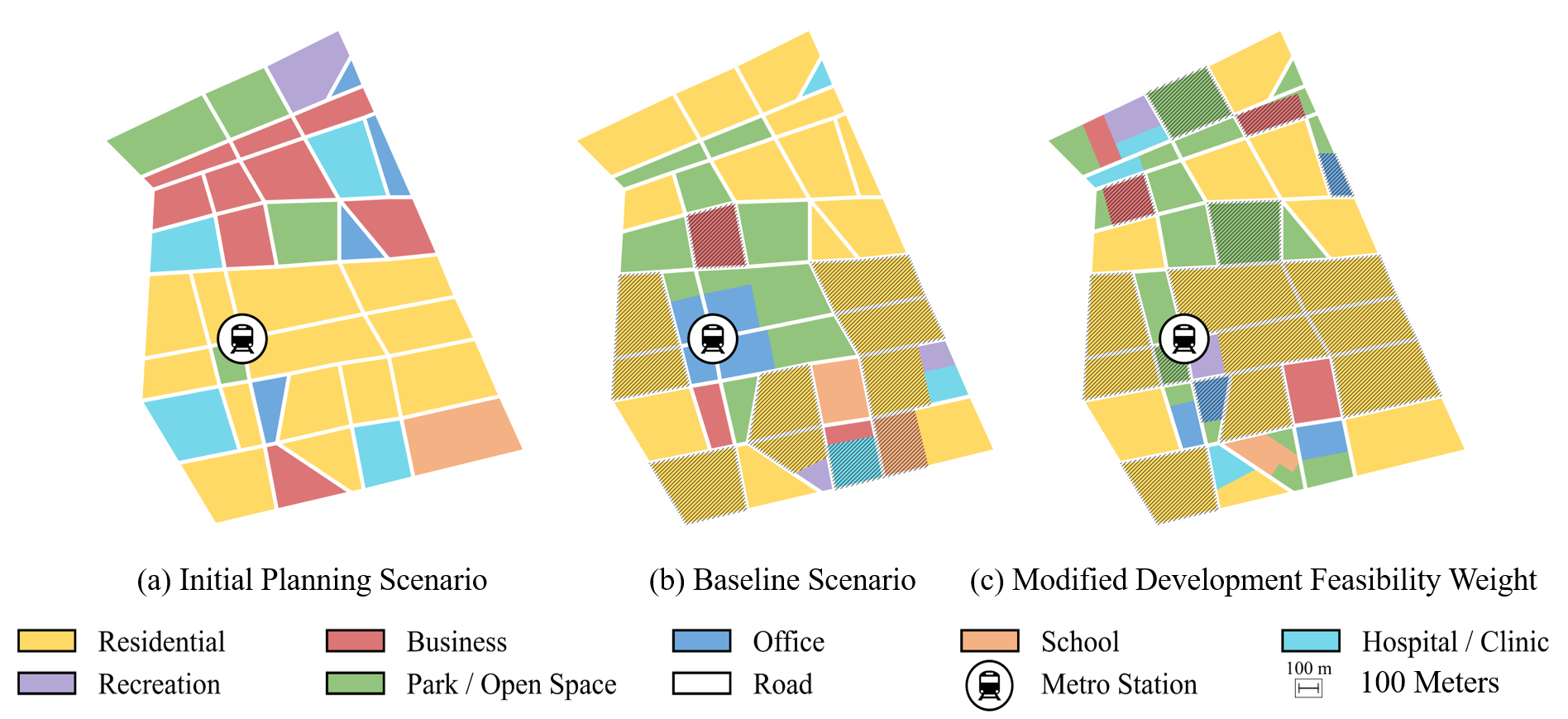}
    \caption{2D spatial planning outcomes under development feasibility weight adjustment. Hatched parcels indicate preserved areas.}
    \label{fig:development_feasibility_2D}
\end{figure}

The spatial outcomes reinforce these numerical patterns. Figures~\ref{fig:development_feasibility_3D} and \ref{fig:development_feasibility_2D} illustrate the 3D and 2D spatial configurations under varying development-feasibility weights, with preserved area indicated by hatched parcels. Compared to the baseline scenario, the modified plan retains a much larger proportion of parcels unchanged from the initial scheme. This reflects a deliberate preservation of existing land-use assignments, highlighting the model's preference for minimizing reallocation in areas where development feasibility is low. While this enhances spatial realism and reduces reconstruction disturbance, it also constrains opportunities for densification and mixed-use clustering around the metro station—both central goals of TOD planning—illustrating the inherent trade-off between achieving high TOD scores and minimizing reconstruction costs.

\subsection{Expert-Informed Application of the Planning Workflow}
\label{subsec:workflow_case_demo}

To demonstrate how human planning judgment is incorporated into the planner-guided workflow shown in Figure~\ref{fig:planning_workflow}, we conduct an expert-informed application for Xintiandi and Jinjing Road Stations. The panel comprises 19 participants with backgrounds in transportation and land-use planning, including students (31.6\%), researchers (36.8\%), and practitioners (26.3\%). A majority (63.2\%) report six or more years of relevant experience, and 94.7\% are currently or have previously been based in Shanghai. The panel participates in two key stages of the workflow: specifying station-specific objective weights to configure the model (Stage~4) and reviewing the resulting candidate plan (Stage~7), thereby demonstrating the process from human-informed model configuration to plan evaluation.

First, they specify the relative importance of LUDE, LUDI, TCU, and DF for each station area. This produces different priority structures for the two cases: the weights for Xintiandi are $(0.23, 0.27, 0.22, 0.28)$, placing relatively greater emphasis on development feasibility and functional diversity in this established central area with strong historical and contextual constraints; the weights for Jinjing Road are $(0.22, 0.30, 0.24, 0.24)$, with land-use diversity receiving the highest priority in its more peripheral and less functionally mixed context. These differences illustrate that TOD priorities are context dependent and that planner knowledge is important in specifying how objectives should be balanced. Using these expert-elicited weights, DRL4TOD and AGA are evaluated under the same spatial inputs, planning constraints, and objective formulation. DRL4TOD outperforms AGA across all reported metrics at both stations (Table~\ref{tab:tod_performance_weight}), indicating that its performance advantage is maintained under independently specified planning priorities.

\begin{table}[!htbp]
  \centering
  \footnotesize
  {
  \caption{Performance of AGA and DRL4TOD under expert-elicited objective weights.}
  \begin{tabular}{c|c|ccccc}
    \toprule
    \multirow{2}{*}{Station} &
    \multirow{2}{*}{Model} &
    \multicolumn{5}{c}{Metrics} \\
    & & Total Reward & LUDE & LUDI & TCU & DF \\
    \midrule
    \multirow{3}{*}{Xintiandi}
      & \textit{Weight}
      & -- & 0.23 & 0.27 & 0.22 & 0.28 \\
      & AGA
      & 0.3865 & 0.1310 & 0.7253 & 0.1716 & 0.4386 \\
      & DRL4TOD
      & \textbf{\underline{0.4349}}$^{**}$ & \textbf{\underline{0.2232}}$^{**}$ & \textbf{\underline{0.8022}}$^{**}$ & \textbf{\underline{0.1859}}$^{**}$ & \textbf{\underline{0.4503}}$^{**}$ \\
    \midrule
    \multirow{3}{*}{Jinjing Road}
      & \textit{Weight}
      & -- & 0.22 & 0.30 & 0.24 & 0.24 \\
      & AGA
      & 0.3354 & 0.2158 & 0.6154 & 0.1680 & 0.2623 \\
      & DRL4TOD
      & \textbf{\underline{0.4153}}$^{**}$ & \textbf{\underline{0.2781}}$^{**}$ & \textbf{\underline{0.7649}}$^{**}$ & \textbf{\underline{0.1963}}$^{**}$ & \textbf{\underline{0.3231}}$^{**}$ \\
    \bottomrule
  \end{tabular}
  \label{tab:tod_performance_weight}
  }
\end{table}

Following candidate generation, the panel reviews the resulting Xintiandi plan against considerations beyond the quantitative objectives. Respondents generally affirm its core TOD outcomes, with density (63.2\%) and design (63.2\%) most frequently identified as strengths, followed by destination accessibility (57.9\%). At the same time, they identify several context-specific issues: some note that the proposed land-use configuration does not sufficiently respond to Xintiandi's central-city functions and distinctive historical and cultural context, making the plan appear similar to a more generic station-area development; others raise pedestrian-scale concerns that concentrated high-intensity office and business development could increase thermal discomfort and reduce shading. These comments illustrate the complementary role of planner review in assessing local identity, urban design quality, and other contextual considerations not fully captured by the quantitative objectives, with the resulting feedback informing revisions to objective weights or planning constraints and subsequent regeneration of the plan.

\section{Conclusion and Discussion}
\label{sec:discussion}

DRL4TOD provides a planner-guided decision-support framework for generating parcel-level transit-oriented land-use alternatives under multiple planning objectives and site-specific constraints. The workflow connects planner-led problem definition, local spatial-data preparation, objective and constraint specification, graph-based optimization, and candidate-plan review. Planners can define the target land-use composition, assign relative importance to land-use density, land-use diversity, transit capacity utilization, and development feasibility, and designate parcels that should remain unchanged. DRL4TOD then translates these inputs into a parcel--road--station optimization problem and generates candidate configurations of land-use types and FAR for the station area. The framework can be configured for different planning contexts through locally harmonized spatial data, planning requirements, objective weights, and, where necessary, locally calibrated components such as the ridership model. The generated alternatives can subsequently be reviewed against qualitative contextual considerations, and the planner can revise the objective weights or planning constraints when further adjustment is required.

Under the same spatial inputs, planning constraints, objective formulation, and evaluation protocol, DRL4TOD achieves the strongest overall performance among the evaluated methods across the two station-area cases (Section~\ref{subsec:performance_comparison}). The resulting configurations generally place higher-intensity employment-related uses closer to the transit node while retaining a diverse mix of complementary land uses within the station catchment. This station-proximate concentration of development intensity is consistent with earlier TOD optimization studies, which associate higher development intensity near transit stations with improved ridership or sustainability outcomes \citep{lin2006tod,li2010transit}. At the same time, the results show that this spatial pattern is conditioned by the broader set of planning objectives. When land-use diversity or transit capacity utilization receives greater emphasis, lower-intensity and complementary uses, including recreational, park, school, and healthcare uses, remain within the station catchment alongside higher-intensity development; when development feasibility is prioritized, a larger share of the existing urban fabric is retained (Section~\ref{subsec:flexibility_analysis}). The resulting plans therefore do not simply maximize FAR around the station. They reflect a joint spatial trade-off among development intensity, functional diversity, transit performance, and redevelopment burden. This finding extends the interpretation of station-proximate density in prior TOD studies by showing how the pattern changes when several planning objectives are optimized simultaneously. The observed performance is associated with the use of station-aware and neighborhood-aware spatial representations: the GNN encodes relationships among parcels, roads, and the metro station, while the reinforcement-learning policy selects land-use and FAR actions sequentially in response to the evolving spatial state. The ablation results support the importance of these graph-based spatial representations for the planning task (Section~\ref{subsec:ablation_analysis}).

The practical value of DRL4TOD lies in making the spatial consequences of planning priorities visible and comparable before a plan is adopted. The planning-priority analysis demonstrates how planners can generate alternatives that place different emphasis on development intensity, land-use diversity, transit capacity utilization, or retention of the existing urban fabric (Section~\ref{subsec:flexibility_analysis}). The preservation-constrained scenario shows how parcels subject to a preservation requirement can be fixed through the planning constraints while the remaining parcels continue to be optimized (Appendix~\ref{appendix:preservation_scenario}). The expert-informed application further demonstrates how human input can enter both model configuration and candidate-plan review: participants specified station-specific objective weights, DRL4TOD generated candidate plans, and the resulting Xintiandi plan was reviewed with respect to contextual and design considerations beyond the quantitative objectives (Section~\ref{subsec:workflow_case_demo}). In this workflow, the model supports the generation and evaluation of spatial alternatives, while planners retain responsibility for interpreting local identity, pedestrian experience, legal conditions, social impacts, and other considerations that are not fully represented by the quantitative objectives (Section~\ref{sec:planning_workflow}). Its practical role is therefore to translate planning priorities into spatially explicit alternatives, allowing planners to examine the consequences of different priorities and constraints before selecting or refining a plan.

The scope of this planner-guided formulation also defines its main limitations and directions for further development. One boundary concerns qualitative, behavioral, and social dimensions of the built environment, such as pedestrian experience and neighborhood character, which are not directly represented in the current reward formulation and therefore remain part of the planner-led review process. Higher density and diversity should likewise not be interpreted as inherently equitable outcomes, since the current formulation does not include measures of affordability, tenure security, or displacement risk. Future versions could incorporate these dimensions as additional reward components or planning constraints using locally available housing, household, and demographic data. A related boundary concerns development feasibility. Demolition and reconstruction volume captures the physical scale of intervention but not the legal, political, and social conditions that determine whether redevelopment can actually be carried out, including property-rights transfer, land-acquisition procedures, inter-agency and community approval, and compensation costs for displaced residents, which may be distinct from and exceed construction costs. A more comprehensive formulation could combine the existing physical proxy with locally available information on redevelopment procedures, institutional constraints, and compensation requirements; explicit preservation or legal requirements could be represented through hard constraints or action masking, while conditions that cannot be reliably quantified would remain subject to planner review. At the representation level, each final parcel is assigned a single land-use type, so mixed-use development within an individual parcel is not represented explicitly. This limitation could be addressed by extending the action space to include primary and secondary uses with corresponding floor-area shares. The model also optimizes a static snapshot of the station area and therefore does not capture long-term changes in urban growth, infrastructure condition, or demographic needs. A multi-period formulation could introduce time-varying state attributes related to building condition, development demand, population structure, transit use, and economic activity. Finally, the empirical evaluation is limited to two station areas in Shanghai; broader transferability should be assessed across additional station types and cities using locally harmonized spatial data, planning requirements, and calibrated ridership models.

\section*{Declaration of Generative AI and AI-assisted technologies in the writing process}
During the preparation of this work the authors used ChatGPT in order to polish the grammar and wording of part of the manuscript. After using this tool/service, the authors reviewed and edited the content as needed and take full responsibility for the content of the publication.

\appendix
\section{The transit capacity utilization simulation model}
\label{appendix:transit}

We develop a station-level ridership prediction model using metro ridership data from 337 stations in the 2018 Shanghai metro system \citep{ding2024graph}. The model uses three categories of built-environment features as inputs: (1) distance-decay-weighted density measures for each land use type, (2) an entropy-based land-use diversity index, and (3) road network accessibility indicators, calculated by averaging the network distances from parcel centroids to the metro station within each land use category. An XGBoost regressor is adopted to capture nonlinear relationships between built-environment characteristics and average daily ridership.

Model performance is evaluated using five repeated random holdout splits. In each repetition, the 337 stations are randomly divided into an 80\% training set (269 stations) and a 20\% test set (68 stations). Across the five test sets, the model achieves an averaged Pearson correlation of 0.591, a mean normalized RMSE of 0.125, and a mean normalized MAE of 0.093. After inverse transformation to the original ridership scale, the corresponding RMSE and MAE are 15{,}601 and 11{,}632 passenger trips per day, respectively.

\begin{figure}[!htbp]
  \centering
  \includegraphics[width=\textwidth]{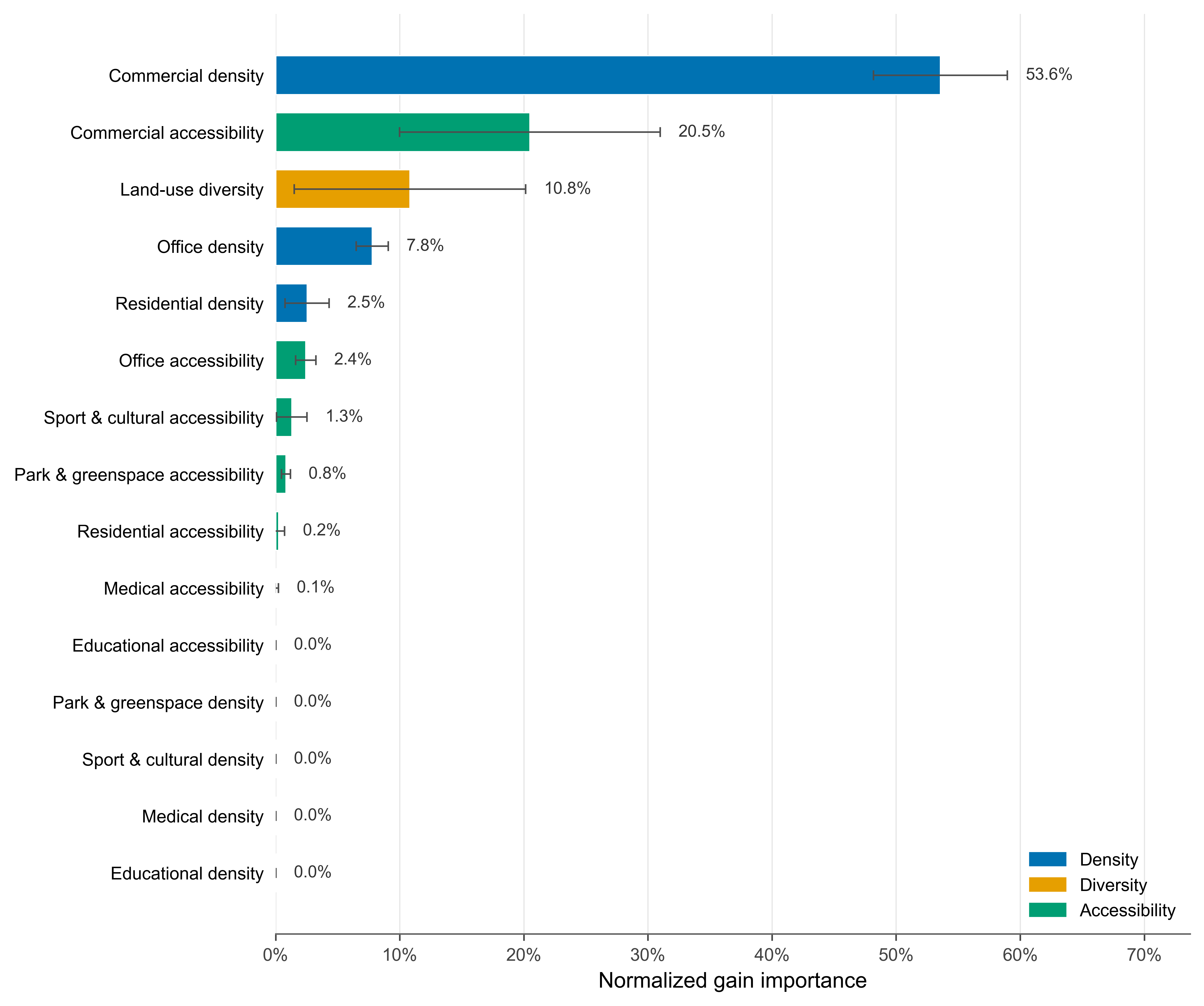}
  \caption{XGBoost feature importance for the station-level ridership model.}
  \label{fig:xgb_feature_importance}
\end{figure}

Feature importance is quantified using normalized XGBoost gain and averaged across the same five random splits, as shown in Figure~\ref{fig:xgb_feature_importance}. Collectively, density, accessibility, and land-use diversity account for 63.9\%, 25.3\%, and 10.8\% of the total feature importance, respectively. At the individual-feature level, commercial density is the dominant predictor (53.6\%), followed by commercial accessibility (20.5\%), land-use diversity (10.8\%), office density (7.8\%), residential density (2.5\%), and office accessibility (2.4\%). The remaining features each contribute less than 1.5\%. The trained model is then embedded within the DRL framework to estimate ridership outcomes under alternative land-use configurations, providing a data-driven signal for optimizing transit capacity utilization.

\section{Architecture and Training Sensitivity}
\label{appendix:sensitivity_analysis}

To assess whether the reported performance depends on a particular network architecture or training setting, we conduct a controlled sensitivity analysis by varying the GNN message-passing depth and the PPO learning rate. The reference configuration uses a learning rate of $4\times10^{-4}$ and two GNN layers. Each experiment varies one parameter while keeping the graph inputs, hidden dimension, reward weights, FAR constraints, training budget, and inference procedure unchanged, and each setting is repeated five times using the same random seeds as the main experiments, with results reported as means. Table~\ref{tab:sensitivity_xintiandi} summarizes the results for the Xintiandi case.

The reference configuration achieves the highest mean Total Reward, with both a lower ($1\times10^{-4}$) and a higher ($1\times10^{-3}$) learning rate reducing performance, plausibly because a lower rate leaves the policy short of convergence within the fixed training budget, while a higher rate introduces a noisier optimization trajectory. GNN depth follows a similar pattern: a single message-passing layer restricts each parcel's representation to its immediate neighbors, limiting the model's ability to capture station-mediated relationships between non-adjacent parcels, while three layers modestly reduce performance relative to the two-layer reference, consistent with over-smoothing, in which repeated aggregation makes node embeddings increasingly indistinguishable from their local neighborhood average.

All tested configurations achieve a higher mean Total Reward than the strongest Xintiandi baseline, AGA, which obtains 0.5826. The improvement ranges from 19.2\% for the one-layer GNN to 78.0\% for the reference configuration. These results indicate that DRL4TOD's advantage is maintained across the tested learning rates and GNN depths rather than arising from a single finely tuned configuration.

\begin{table}[!htbp]
  \centering
  \footnotesize
  {
  \caption{Sensitivity of DRL4TOD performance to the PPO learning rate and GNN message-passing depth at Xintiandi Station. Values are means over five independent runs. In each block, only the listed parameter is varied; all other settings remain fixed. Bold and underlining identify the reference configuration.}
  \label{tab:sensitivity_xintiandi}
  \begin{tabular}{l c rrrrr}
    \toprule
    \multirow{2}{*}{Hyperparameter} & \multirow{2}{*}{Value} &
    \multicolumn{5}{c}{Metrics} \\
    & & Total Reward & LUDE & LUDI & TCU & DF \\
    \midrule
    \multirow{3}{*}{PPO learning rate}
      & $1\times10^{-4}$ & 0.7756 & 1.1848 & 0.8217 & 0.2107 & 0.4472 \\
      & $4\times10^{-4}$ &
        \textbf{\underline{1.0371}} &
        \textbf{\underline{1.6965}} &
        \textbf{\underline{0.8385}} &
        \textbf{\underline{0.2272}} &
        \textbf{\underline{0.4505}} \\
      & $1\times10^{-3}$ & 0.9548 & 1.5478 & 0.8142 & 0.2121 & 0.4359 \\
    \midrule
    \multirow{3}{*}{GNN layers}
      & 1 & 0.6946 & 1.0332 & 0.8082 & 0.2098 & 0.4258 \\
      & 2 &
        \textbf{\underline{1.0371}} &
        \textbf{\underline{1.6965}} &
        \textbf{\underline{0.8385}} &
        \textbf{\underline{0.2272}} &
        \textbf{\underline{0.4505}} \\
      & 3 & 0.9407 & 1.5163 & 0.8291 & 0.2056 & 0.4487 \\
    \bottomrule
  \end{tabular}
  }
\end{table}

\section{Preservation-Constrained Scenario}
\label{appendix:preservation_scenario}

The preceding experiments allow all parcels within the station area to be reassigned. To examine whether DRL4TOD can accommodate hard redevelopment constraints, we introduce a preservation-constrained scenario at Xintiandi Station. Parcels containing preserved shikumen heritage buildings are designated as non-redevelopable and retain their existing land-use types and FAR values throughout the optimization. These parcels remain in the graph representation but are excluded from the feasible action space, while all other parcels remain available for reassignment. All other model settings are unchanged from the unconstrained experiment.

\begin{figure}[!htb]
  \centering
  \includegraphics[width=\textwidth]{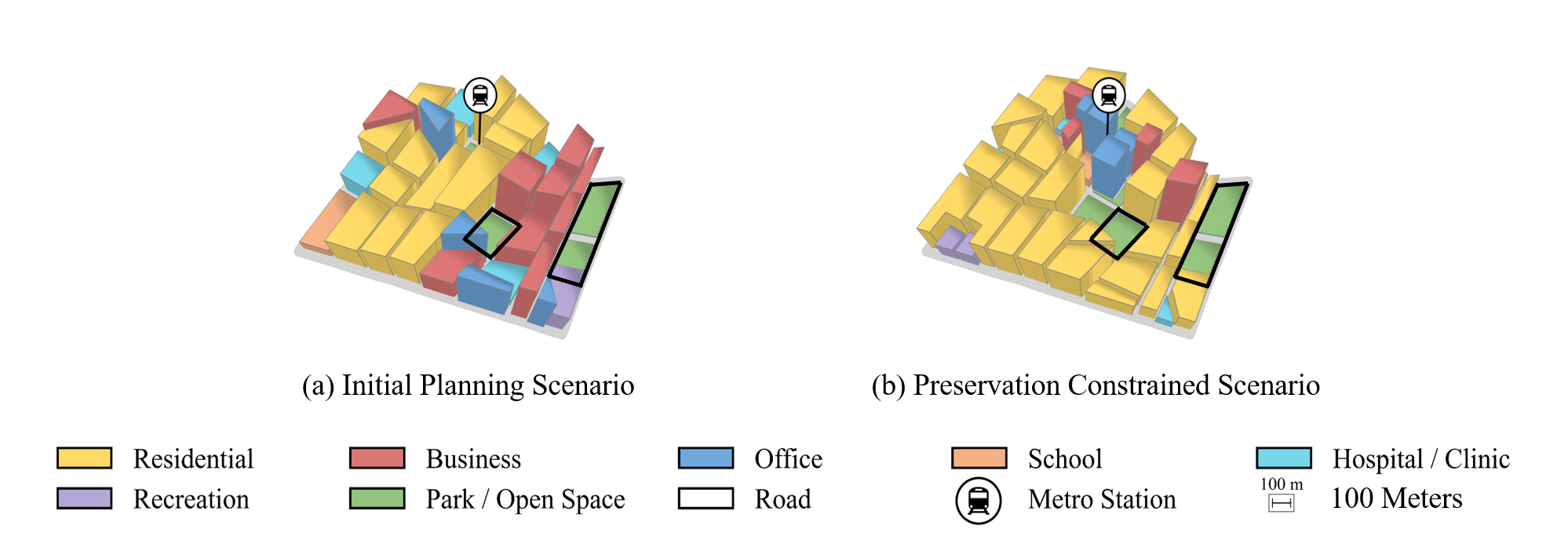}
  \vspace{-20pt}
  \caption{3D visualization of the preservation-constrained scenario at Xintiandi Station: (a) initial planning configuration and (b) preservation-constrained configuration. Solid outlines in both panels indicate the parcels fixed in the preservation-constrained scenario.}
  \label{fig:preservation_3d_results}
\end{figure}

As shown in Table~\ref{tab:preservation_performance}, restricting the feasible action space reduces the achievable performance, with Total Reward decreasing from 1.0371 in the unconstrained case to 0.6889. Nevertheless, the constrained solution remains substantially better than the Initial Plan (0.1757), indicating that meaningful TOD-oriented improvements can still be achieved while respecting non-redevelopable heritage parcels. Figure~\ref{fig:preservation_3d_results} further shows that the designated heritage parcels remain unchanged, while the model reorganizes the remaining developable parcels and retains a broadly similar transit-oriented spatial structure to the unconstrained configuration.

\begin{table}[!htbp]
  \centering
  \footnotesize
  {
  \caption{Performance comparison under heritage-preservation constraints at Xintiandi Station.}
  \label{tab:preservation_performance}
  \begin{tabular}{c|ccccc}
    \toprule
    \multirow{2}{*}{Model Variant} & \multicolumn{5}{c}{Performance Metrics} \\
    & Total Reward & LUDE & LUDI & TCU & DF \\
    \midrule
    Initial Plan & 0.1757 & 0.1732 & 0.7858 & 0.1993 & - \\
    DRL4TOD & 1.0371 & 1.6965 & 0.8385 & 0.2272 & 0.4505 \\
    Preservation-Constrained Scenario & 0.6889 & 1.0002 & 0.7313 & 0.2598 & 0.4362 \\
    \bottomrule
  \end{tabular}
  }
\end{table}

\bibliographystyle{model5-names2}\biboptions{authoryear}
\bibliography{ref}

\begin{thebibliography}{54}
\expandafter\ifx\csname natexlab\endcsname\relax\def\natexlab#1{#1}\fi
\providecommand{\url}[1]{\texttt{#1}}
\providecommand{\href}[2]{#2}
\providecommand{\path}[1]{#1}
\providecommand{\DOIprefix}{doi:}
\providecommand{\ArXivprefix}{arXiv:}
\providecommand{\URLprefix}{URL: }
\providecommand{\Pubmedprefix}{pmid:}
\providecommand{\doi}[1]{\href{http://dx.doi.org/#1}{\path{#1}}}
\providecommand{\Pubmed}[1]{\href{pmid:#1}{\path{#1}}}
\providecommand{\bibinfo}[2]{#2}
\ifx\xfnm\relax \def\xfnm[#1]{\unskip,\space#1}\fi
\bibitem[{Abhijna et~al.(2025)Abhijna, Rao and Perumal}]{abhijna2025new}
\bibinfo{author}{Abhijna, M.}, \bibinfo{author}{Rao, K.~K.}, and
  \bibinfo{author}{Perumal, V.} (\bibinfo{year}{2025}).
\newblock \bibinfo{title}{How new metro lines shape a sustainable future: A
  before-after study of travel behavior, perceptions, and emissions in mumbai
  metropolitan region, india}.
\newblock {\it \bibinfo{journal}{Sustainable Cities and Society}\/},  {\it
  \bibinfo{volume}{127}\/}, \bibinfo{pages}{106431}.
\bibitem[{Ashik et~al.(2022)Ashik, Rahman and
  Kamruzzaman}]{ashik2022investigating}
\bibinfo{author}{Ashik, F.~R.}, \bibinfo{author}{Rahman, M.}, and
  \bibinfo{author}{Kamruzzaman, M.} (\bibinfo{year}{2022}).
\newblock \bibinfo{title}{Investigating the impacts of transit-oriented
  development on transport-related co2 emissions}.
\newblock {\it \bibinfo{journal}{Transportation Research Part D: Transport and
  Environment}\/},  {\it \bibinfo{volume}{105}\/}, \bibinfo{pages}{103227}.
\bibitem[{Banai(1998)}]{banai1998transit}
\bibinfo{author}{Banai, R.} (\bibinfo{year}{1998}).
\newblock \bibinfo{title}{Transit-oriented development suitability analysis by
  the analytic hierarchy process and a geographic information system: A
  prototype procedure}.
\newblock {\it \bibinfo{journal}{Journal of Public Transportation}\/},  {\it
  \bibinfo{volume}{2}\/}, \bibinfo{pages}{43--65}.
\bibitem[{Bertolini(1999)}]{bertolini1999spatial}
\bibinfo{author}{Bertolini, L.} (\bibinfo{year}{1999}).
\newblock \bibinfo{title}{Spatial development patterns and public transport:
  the application of an analytical model in the netherlands}.
\newblock {\it \bibinfo{journal}{Planning Practice and Research}\/},  {\it
  \bibinfo{volume}{14}\/}, \bibinfo{pages}{199--210}.
\bibitem[{Bertsekas(2012)}]{bertsekas2012dynamic}
\bibinfo{author}{Bertsekas, D.} (\bibinfo{year}{2012}).
\newblock {\it \bibinfo{title}{Dynamic programming and optimal control: Volume
  I}\/} volume~\bibinfo{volume}{4}.
\newblock \bibinfo{publisher}{Athena scientific}.
\bibitem[{Calthorpe(1993)}]{calthorpe1993next}
\bibinfo{author}{Calthorpe, P.} (\bibinfo{year}{1993}).
\newblock {\it \bibinfo{title}{The next American metropolis}\/}
  volume~\bibinfo{volume}{23}.
\newblock \bibinfo{publisher}{New York: Princeton Architectural Press}.
\bibitem[{Cervero(2007)}]{cervero2007transit}
\bibinfo{author}{Cervero, R.} (\bibinfo{year}{2007}).
\newblock \bibinfo{title}{Transit-oriented development's ridership bonus: a
  product of self-selection and public policies}.
\newblock {\it \bibinfo{journal}{Environment and Planning A}\/},  {\it
  \bibinfo{volume}{39}\/}, \bibinfo{pages}{2068--2085}.
\bibitem[{Cervero and Kockelman(1997)}]{cervero1997travel}
\bibinfo{author}{Cervero, R.}, and \bibinfo{author}{Kockelman, K.}
  (\bibinfo{year}{1997}).
\newblock \bibinfo{title}{Travel demand and the 3ds: Density, diversity, and
  design}.
\newblock {\it \bibinfo{journal}{Transportation Research Part D: Transport and
  Environment}\/},  {\it \bibinfo{volume}{2}\/}, \bibinfo{pages}{199--219}.
\bibitem[{Cucuzzella et~al.(2022)Cucuzzella, Owen, Goubran and
  Walker}]{cucuzzella2022tod}
\bibinfo{author}{Cucuzzella, C.}, \bibinfo{author}{Owen, J.},
  \bibinfo{author}{Goubran, S.}, and \bibinfo{author}{Walker, T.}
  (\bibinfo{year}{2022}).
\newblock \bibinfo{title}{A tod index integrating development potential,
  economic vibrancy, and socio-economic factors for encouraging polycentric
  cities}.
\newblock {\it \bibinfo{journal}{Cities}\/},  {\it \bibinfo{volume}{131}\/},
  \bibinfo{pages}{103980}.
\bibitem[{Ding et~al.(2022)Ding, Song, Qiao, Zhang, Yu and
  Zhang}]{ding2022integrating}
\bibinfo{author}{Ding, C.}, \bibinfo{author}{Song, Y.}, \bibinfo{author}{Qiao,
  X.}, \bibinfo{author}{Zhang, Y.}, \bibinfo{author}{Yu, B.}, and
  \bibinfo{author}{Zhang, X.} (\bibinfo{year}{2022}).
\newblock \bibinfo{title}{Integrating land use and greenway network for metro
  station area transit-oriented development}.
\newblock {\it \bibinfo{journal}{Transportation Research Part D: Transport and
  Environment}\/},  {\it \bibinfo{volume}{113}\/}, \bibinfo{pages}{103539}.
\bibitem[{Ding et~al.(2024)Ding, Liang, Wang, Tang, Zhou and
  Zhao}]{ding2024graph}
\bibinfo{author}{Ding, F.}, \bibinfo{author}{Liang, Y.}, \bibinfo{author}{Wang,
  Y.}, \bibinfo{author}{Tang, Y.}, \bibinfo{author}{Zhou, Y.}, and
  \bibinfo{author}{Zhao, Z.} (\bibinfo{year}{2024}).
\newblock \bibinfo{title}{A graph deep learning model for station ridership
  prediction in expanding metro networks}.
\newblock In {\it \bibinfo{booktitle}{Proceedings of the 2nd ACM SIGSPATIAL
  International Workshop on Advances in Urban-AI}\/} (pp.
  \bibinfo{pages}{6--14}).
\bibitem[{Ewing and Cervero(2001)}]{ewing2001travel}
\bibinfo{author}{Ewing, R.}, and \bibinfo{author}{Cervero, R.}
  (\bibinfo{year}{2001}).
\newblock \bibinfo{title}{Travel and the built environment: a synthesis}.
\newblock {\it \bibinfo{journal}{Transportation Research Record}\/},  {\it
  \bibinfo{volume}{1780}\/}, \bibinfo{pages}{87--114}.
\bibitem[{Ewing and Cervero(2010)}]{ewing2010travel}
\bibinfo{author}{Ewing, R.}, and \bibinfo{author}{Cervero, R.}
  (\bibinfo{year}{2010}).
\newblock \bibinfo{title}{Travel and the built environment: A meta-analysis}.
\newblock {\it \bibinfo{journal}{Journal of the American planning
  association}\/},  {\it \bibinfo{volume}{76}\/}, \bibinfo{pages}{265--294}.
\bibitem[{Gao et~al.(2021)Gao, Wang, Cushman, Cheng, Song and
  Ye}]{gao2021sustainable}
\bibinfo{author}{Gao, P.}, \bibinfo{author}{Wang, H.},
  \bibinfo{author}{Cushman, S.~A.}, \bibinfo{author}{Cheng, C.},
  \bibinfo{author}{Song, C.}, and \bibinfo{author}{Ye, S.}
  (\bibinfo{year}{2021}).
\newblock \bibinfo{title}{Sustainable land-use optimization using nsga-ii:
  Theoretical and experimental comparisons of improved algorithms}.
\newblock {\it \bibinfo{journal}{Landscape Ecology}\/},  {\it
  \bibinfo{volume}{36}\/}, \bibinfo{pages}{1877--1892}.
\bibitem[{{Google Research}(2025)}]{GoogleOpenBuildings}
\bibinfo{author}{{Google Research}} (\bibinfo{year}{2025}).
\newblock \bibinfo{title}{Open buildings}.
\newblock \URLprefix \url{https://sites.research.google/gr/open-buildings/}
  \bibinfo{note}{accessed: March 31, 2025}.
\bibitem[{Hosseini et~al.(2025)Hosseini, Atazadeh and
  Rajabifard}]{hosseini2025towards}
\bibinfo{author}{Hosseini, H.}, \bibinfo{author}{Atazadeh, B.}, and
  \bibinfo{author}{Rajabifard, A.} (\bibinfo{year}{2025}).
\newblock \bibinfo{title}{Towards intelligent land administration systems:
  Research challenges, applications and prospects in ai-driven approaches}.
\newblock {\it \bibinfo{journal}{Land Use Policy}\/},  {\it
  \bibinfo{volume}{157}\/}, \bibinfo{pages}{107652}.
\bibitem[{Ibraeva et~al.(2020)Ibraeva, de~Almeida~Correia, Silva and
  Antunes}]{ibraeva2020transit}
\bibinfo{author}{Ibraeva, A.}, \bibinfo{author}{de~Almeida~Correia, G.~H.},
  \bibinfo{author}{Silva, C.}, and \bibinfo{author}{Antunes, A.~P.}
  (\bibinfo{year}{2020}).
\newblock \bibinfo{title}{Transit-oriented development: A review of research
  achievements and challenges}.
\newblock {\it \bibinfo{journal}{Transportation Research Part A: Policy and
  Practice}\/},  {\it \bibinfo{volume}{132}\/}, \bibinfo{pages}{110--130}.
\bibitem[{Ibraeva et~al.(2023)Ibraeva, Correia, Silva and
  Pais~Antunes}]{ibraeva2023impacts}
\bibinfo{author}{Ibraeva, A.}, \bibinfo{author}{Correia, G. H. d.~A.},
  \bibinfo{author}{Silva, C.}, and \bibinfo{author}{Pais~Antunes, A.}
  (\bibinfo{year}{2023}).
\newblock \bibinfo{title}{Impacts of transit-oriented development on car use
  over a 10-year period in porto, portugal: From macro-to micro-analysis}.
\newblock {\it \bibinfo{journal}{International Journal of Sustainable
  Transportation}\/},  {\it \bibinfo{volume}{17}\/},
  \bibinfo{pages}{1195--1206}.
\bibitem[{Kennedy and Eberhart(1995)}]{kennedy1995particle}
\bibinfo{author}{Kennedy, J.}, and \bibinfo{author}{Eberhart, R.}
  (\bibinfo{year}{1995}).
\newblock \bibinfo{title}{Particle swarm optimization}.
\newblock In {\it \bibinfo{booktitle}{Proceedings of ICNN'95-international
  conference on neural networks}\/} (pp. \bibinfo{pages}{1942--1948}).
\newblock \bibinfo{organization}{ieee} volume~\bibinfo{volume}{4}.
\bibitem[{Kirkpatrick et~al.(1983)Kirkpatrick, Gelatt~Jr and
  Vecchi}]{kirkpatrick1983optimization}
\bibinfo{author}{Kirkpatrick, S.}, \bibinfo{author}{Gelatt~Jr, C.~D.}, and
  \bibinfo{author}{Vecchi, M.~P.} (\bibinfo{year}{1983}).
\newblock \bibinfo{title}{Optimization by simulated annealing}.
\newblock {\it \bibinfo{journal}{science}\/},  {\it \bibinfo{volume}{220}\/},
  \bibinfo{pages}{671--680}.
\bibitem[{Kong et~al.(2025)Kong, Feng and Liu}]{kong2025underground}
\bibinfo{author}{Kong, W.}, \bibinfo{author}{Feng, W.}, and
  \bibinfo{author}{Liu, Y.} (\bibinfo{year}{2025}).
\newblock \bibinfo{title}{Underground space planning optimization under the tod
  model using nsga-ii: A case study of qingdaobei railway station and its
  surroundings}.
\newblock {\it \bibinfo{journal}{Sustainability}\/},  {\it
  \bibinfo{volume}{17}\/}, \bibinfo{pages}{9761}.
\bibitem[{Kunz(2010)}]{kunz2010analytic}
\bibinfo{author}{Kunz, J.} (\bibinfo{year}{2010}).
\newblock \bibinfo{title}{The analytic hierarchy process (ahp)}.
\newblock {\it \bibinfo{journal}{Eagle City Hall Location Options Task
  Force}\/},  (pp. \bibinfo{pages}{1--25}).
\bibitem[{Li and Ma(2018)}]{li2018improved}
\bibinfo{author}{Li, X.}, and \bibinfo{author}{Ma, X.} (\bibinfo{year}{2018}).
\newblock \bibinfo{title}{An improved simulated annealing algorithm for
  interactive multi-objective land resource spatial allocation}.
\newblock {\it \bibinfo{journal}{Ecological Complexity}\/},  {\it
  \bibinfo{volume}{36}\/}, \bibinfo{pages}{184--195}.
\bibitem[{Li et~al.(2010)Li, Guo, Li, Xu, Wang and Kong}]{li2010transit}
\bibinfo{author}{Li, Y.}, \bibinfo{author}{Guo, H.}, \bibinfo{author}{Li, H.},
  \bibinfo{author}{Xu, G.}, \bibinfo{author}{Wang, Z.}, and
  \bibinfo{author}{Kong, C.} (\bibinfo{year}{2010}).
\newblock \bibinfo{title}{Transit-oriented land planning model considering
  sustainability of mass rail transit}.
\newblock {\it \bibinfo{journal}{Journal of Urban Planning and Development}\/},
   {\it \bibinfo{volume}{136}\/}, \bibinfo{pages}{243--248}.
\bibitem[{Li et~al.(2019)Li, Han, Xin, Luo, Su and Weng}]{li2019transit}
\bibinfo{author}{Li, Z.}, \bibinfo{author}{Han, Z.}, \bibinfo{author}{Xin, J.},
  \bibinfo{author}{Luo, X.}, \bibinfo{author}{Su, S.}, and
  \bibinfo{author}{Weng, M.} (\bibinfo{year}{2019}).
\newblock \bibinfo{title}{Transit oriented development among metro station
  areas in shanghai, china: Variations, typology, optimization and implications
  for land use planning}.
\newblock {\it \bibinfo{journal}{Land use policy}\/},  {\it
  \bibinfo{volume}{82}\/}, \bibinfo{pages}{269--282}.
\bibitem[{Liang et~al.(2024)Liang, Wang, Li, Zhou, Chen, Zhang and
  Chen}]{liang2024sponet}
\bibinfo{author}{Liang, H.}, \bibinfo{author}{Wang, S.}, \bibinfo{author}{Li,
  H.}, \bibinfo{author}{Zhou, L.}, \bibinfo{author}{Chen, H.},
  \bibinfo{author}{Zhang, X.}, and \bibinfo{author}{Chen, X.}
  (\bibinfo{year}{2024}).
\newblock \bibinfo{title}{Sponet: solve spatial optimization problem using deep
  reinforcement learning for urban spatial decision analysis}.
\newblock {\it \bibinfo{journal}{International Journal of Digital Earth}\/},
  {\it \bibinfo{volume}{17}\/}, \bibinfo{pages}{2299211}.
\bibitem[{Lin and Gau(2006)}]{lin2006tod}
\bibinfo{author}{Lin, J.-J.}, and \bibinfo{author}{Gau, C.}
  (\bibinfo{year}{2006}).
\newblock \bibinfo{title}{A tod planning model to review the regulation of
  allowable development densities around subway stations}.
\newblock {\it \bibinfo{journal}{Land Use Policy}\/},  {\it
  \bibinfo{volume}{23}\/}, \bibinfo{pages}{353--360}.
\bibitem[{Liu et~al.(2020)Liu, Zhang and Xu}]{liu2020conceptual}
\bibinfo{author}{Liu, L.}, \bibinfo{author}{Zhang, M.}, and
  \bibinfo{author}{Xu, T.} (\bibinfo{year}{2020}).
\newblock \bibinfo{title}{A conceptual framework and implementation tool for
  land use planning for corridor transit oriented development}.
\newblock {\it \bibinfo{journal}{Cities}\/},  {\it \bibinfo{volume}{107}\/},
  \bibinfo{pages}{102939}.
\bibitem[{Liu et~al.(2024)Liu, Chen and Pan}]{liu2024station}
\bibinfo{author}{Liu, X.}, \bibinfo{author}{Chen, X.}, and
  \bibinfo{author}{Pan, H.} (\bibinfo{year}{2024}).
\newblock \bibinfo{title}{From “station-city integration” to “corridor
  integration”: Tod framework and model optimization from the perspective of
  space of flows}.
\newblock In {\it \bibinfo{booktitle}{Urban Plan. Forum}\/} (pp.
  \bibinfo{pages}{34--40}).
\newblock volume~\bibinfo{volume}{4}.
\bibitem[{Ma et~al.(2018)Ma, Chen, Li, Ding and Wang}]{ma2018sustainable}
\bibinfo{author}{Ma, X.}, \bibinfo{author}{Chen, X.}, \bibinfo{author}{Li, X.},
  \bibinfo{author}{Ding, C.}, and \bibinfo{author}{Wang, Y.}
  (\bibinfo{year}{2018}).
\newblock \bibinfo{title}{Sustainable station-level planning: An integrated
  transport and land use design model for transit-oriented development}.
\newblock {\it \bibinfo{journal}{Journal of Cleaner Production}\/},  {\it
  \bibinfo{volume}{170}\/}, \bibinfo{pages}{1052--1063}.
\bibitem[{Mangu et~al.(2025)Mangu, Kadali, Subbarao and
  Lin}]{mangu2025evaluation}
\bibinfo{author}{Mangu, S.}, \bibinfo{author}{Kadali, B.~R.},
  \bibinfo{author}{Subbarao, S.~S.}, and \bibinfo{author}{Lin, J.-J.}
  (\bibinfo{year}{2025}).
\newblock \bibinfo{title}{Evaluation of transit-oriented development based on
  9d’s approach in developing countries context}.
\newblock {\it \bibinfo{journal}{Transport Policy}\/},  {\it
  \bibinfo{volume}{163}\/}, \bibinfo{pages}{138--151}.
\bibitem[{{Microsoft}(2025)}]{MicrosoftBuildingFootprints}
\bibinfo{author}{{Microsoft}} (\bibinfo{year}{2025}).
\newblock \bibinfo{title}{Global ml building footprints}.
\newblock \URLprefix
  \url{https://github.com/microsoft/GlobalMLBuildingFootprints}
  \bibinfo{note}{accessed: March 31, 2025}.
\bibitem[{Okeke et~al.(2020)Okeke, Okosun, Udeh and
  Okekeogbu}]{okeke2020cities}
\bibinfo{author}{Okeke, F.~O.}, \bibinfo{author}{Okosun, A.~E.},
  \bibinfo{author}{Udeh, C.~A.}, and \bibinfo{author}{Okekeogbu, C.~J.}
  (\bibinfo{year}{2020}).
\newblock \bibinfo{title}{Cities for people: the dependency \& impact of
  automobile in the life of city dwellers}.
\newblock {\it \bibinfo{journal}{European Journal of Sustainable
  Development}\/},  {\it \bibinfo{volume}{9}\/}, \bibinfo{pages}{157--157}.
\bibitem[{{OpenStreetMap Contributors}(2025)}]{OpenStreetMap}
\bibinfo{author}{{OpenStreetMap Contributors}} (\bibinfo{year}{2025}).
\newblock \bibinfo{title}{Openstreetmap}.
\newblock \URLprefix \url{https://www.openstreetmap.org}
  \bibinfo{note}{accessed: March 31, 2025}.
\bibitem[{Papa et~al.(2008)Papa, Pagliara and Bertolini}]{papa2008rail}
\bibinfo{author}{Papa, E.}, \bibinfo{author}{Pagliara, F.}, and
  \bibinfo{author}{Bertolini, L.} (\bibinfo{year}{2008}).
\newblock \bibinfo{title}{Rail system development and urban transformations:
  Towards a spatial decision support system}.
\newblock {\it \bibinfo{journal}{Railway development: Impacts on urban
  dynamics}\/},  (pp. \bibinfo{pages}{337--357}).
\bibitem[{Reusser et~al.(2008)Reusser, Loukopoulos, Stauffacher and
  Scholz}]{reusser2008classifying}
\bibinfo{author}{Reusser, D.~E.}, \bibinfo{author}{Loukopoulos, P.},
  \bibinfo{author}{Stauffacher, M.}, and \bibinfo{author}{Scholz, R.~W.}
  (\bibinfo{year}{2008}).
\newblock \bibinfo{title}{Classifying railway stations for sustainable
  transitions--balancing node and place functions}.
\newblock {\it \bibinfo{journal}{Journal of Transport Geography}\/},  {\it
  \bibinfo{volume}{16}\/}, \bibinfo{pages}{191--202}.
\bibitem[{Sahu(2018)}]{sahu2018methodology}
\bibinfo{author}{Sahu, A.} (\bibinfo{year}{2018}).
\newblock \bibinfo{title}{A methodology to modify land uses in a transit
  oriented development scenario}.
\newblock {\it \bibinfo{journal}{Journal of Environmental Management}\/},  {\it
  \bibinfo{volume}{213}\/}, \bibinfo{pages}{467--477}.
\bibitem[{{Shanghai Municipal People's Government}(2020)}]{ShanghaiGov2020}
\bibinfo{author}{{Shanghai Municipal People's Government}}
  (\bibinfo{year}{2020}).
\newblock \bibinfo{title}{Shanghai master plan (2017-2035)}.
\newblock \URLprefix
  \url{https://www.shanghai.gov.cn/nw11056/20200813/0001-11056_859.html}
  \bibinfo{note}{accessed: March 31, 2025}.
\bibitem[{Singh et~al.(2017)Singh, Lukman, Flacke, Zuidgeest and
  Van~Maarseveen}]{singh2017measuring}
\bibinfo{author}{Singh, Y.~J.}, \bibinfo{author}{Lukman, A.},
  \bibinfo{author}{Flacke, J.}, \bibinfo{author}{Zuidgeest, M.}, and
  \bibinfo{author}{Van~Maarseveen, M.} (\bibinfo{year}{2017}).
\newblock \bibinfo{title}{Measuring tod around transit nodes-towards tod
  policy}.
\newblock {\it \bibinfo{journal}{Transport policy}\/},  {\it
  \bibinfo{volume}{56}\/}, \bibinfo{pages}{96--111}.
\bibitem[{Srinivas and Patnaik(2002)}]{srinivas2002adaptive}
\bibinfo{author}{Srinivas, M.}, and \bibinfo{author}{Patnaik, L.~M.}
  (\bibinfo{year}{2002}).
\newblock \bibinfo{title}{Adaptive probabilities of crossover and mutation in
  genetic algorithms}.
\newblock {\it \bibinfo{journal}{IEEE Transactions on Systems, Man, and
  Cybernetics}\/},  {\it \bibinfo{volume}{24}\/}, \bibinfo{pages}{656--667}.
\bibitem[{Su et~al.(2024)Su, Zheng, Ding, Jin and Li}]{su2024metrognn}
\bibinfo{author}{Su, H.}, \bibinfo{author}{Zheng, Y.}, \bibinfo{author}{Ding,
  J.}, \bibinfo{author}{Jin, D.}, and \bibinfo{author}{Li, Y.}
  (\bibinfo{year}{2024}).
\newblock \bibinfo{title}{Metrognn: Metro network expansion with reinforcement
  learning}.
\newblock In {\it \bibinfo{booktitle}{Companion Proceedings of the ACM on Web
  Conference 2024}\/} (pp. \bibinfo{pages}{650--653}).
\bibitem[{Su et~al.(2022)Su, Wang, Li and Kang}]{su2022deciphering}
\bibinfo{author}{Su, S.}, \bibinfo{author}{Wang, Z.}, \bibinfo{author}{Li, B.},
  and \bibinfo{author}{Kang, M.} (\bibinfo{year}{2022}).
\newblock \bibinfo{title}{Deciphering the influence of tod on metro ridership:
  An integrated approach of extended node-place model and interpretable machine
  learning with planning implications}.
\newblock {\it \bibinfo{journal}{Journal of Transport Geography}\/},  {\it
  \bibinfo{volume}{104}\/}, \bibinfo{pages}{103455}.
\bibitem[{Su et~al.(2021)Su, Zhang, He, Zhang, Hu and Kang}]{su2021unraveling}
\bibinfo{author}{Su, S.}, \bibinfo{author}{Zhang, J.}, \bibinfo{author}{He,
  S.}, \bibinfo{author}{Zhang, H.}, \bibinfo{author}{Hu, L.}, and
  \bibinfo{author}{Kang, M.} (\bibinfo{year}{2021}).
\newblock \bibinfo{title}{Unraveling the impact of tod on housing rental prices
  and implications on spatial planning: A comparative analysis of five chinese
  megacities}.
\newblock {\it \bibinfo{journal}{Habitat International}\/},  {\it
  \bibinfo{volume}{107}\/}, \bibinfo{pages}{102309}.
\bibitem[{Sutton(2018)}]{sutton2018reinforcement}
\bibinfo{author}{Sutton, R.~S.} (\bibinfo{year}{2018}).
\newblock \bibinfo{title}{Reinforcement learning: An introduction}.
\newblock {\it \bibinfo{journal}{A Bradford Book}\/}, .
\bibitem[{Thomas and Bertolini(2017)}]{thomas2017defining}
\bibinfo{author}{Thomas, R.}, and \bibinfo{author}{Bertolini, L.}
  (\bibinfo{year}{2017}).
\newblock \bibinfo{title}{Defining critical success factors in tod
  implementation using rough set analysis}.
\newblock {\it \bibinfo{journal}{Journal of Transport and Land Use}\/},  {\it
  \bibinfo{volume}{10}\/}, \bibinfo{pages}{139--154}.
\bibitem[{Vale(2015)}]{vale2015transit}
\bibinfo{author}{Vale, D.~S.} (\bibinfo{year}{2015}).
\newblock \bibinfo{title}{Transit-oriented development, integration of land use
  and transport, and pedestrian accessibility: Combining node-place model with
  pedestrian shed ratio to evaluate and classify station areas in lisbon}.
\newblock {\it \bibinfo{journal}{Journal of Transport Geography}\/},  {\it
  \bibinfo{volume}{45}\/}, \bibinfo{pages}{70--80}.
\bibitem[{Wang et~al.(2026)Wang, Zhou and Cao}]{wang2026multi}
\bibinfo{author}{Wang, Y.}, \bibinfo{author}{Zhou, X.}, and
  \bibinfo{author}{Cao, K.} (\bibinfo{year}{2026}).
\newblock \bibinfo{title}{Multi-objective land use optimization for enhanced
  urban livability using a nsga-iii-differential evolution algorithm}.
\newblock {\it \bibinfo{journal}{Land Use Policy}\/},  {\it
  \bibinfo{volume}{161}\/}, \bibinfo{pages}{107880}.
\bibitem[{Widita et~al.(2025)Widita, Ikaputra and Widyastuti}]{widita2025tod}
\bibinfo{author}{Widita, A.}, \bibinfo{author}{Ikaputra}, and
  \bibinfo{author}{Widyastuti, D.~T.} (\bibinfo{year}{2025}).
\newblock \bibinfo{title}{Tod-related features and station-level ridership:
  insights from the jakarta metropolitan area, indonesia: A. widita et al.}
\newblock {\it \bibinfo{journal}{Public Transport}\/},  {\it
  \bibinfo{volume}{17}\/}, \bibinfo{pages}{505--527}.
\bibitem[{Xiao et~al.(2021)Xiao, Yang and Chi}]{xiao2021subway}
\bibinfo{author}{Xiao, C.}, \bibinfo{author}{Yang, Y.}, and
  \bibinfo{author}{Chi, G.} (\bibinfo{year}{2021}).
\newblock \bibinfo{title}{Subway development and obesity: Evidence from china}.
\newblock {\it \bibinfo{journal}{Journal of Transport \& Health}\/},  {\it
  \bibinfo{volume}{21}\/}, \bibinfo{pages}{101065}.
\bibitem[{Zhao et~al.(2025)Zhao, Shen and Zhao}]{zhao2025large}
\bibinfo{author}{Zhao, L.}, \bibinfo{author}{Shen, S.}, and
  \bibinfo{author}{Zhao, Z.} (\bibinfo{year}{2025}).
\newblock \bibinfo{title}{Large-scale electric bus network transition planning
  via deep reinforcement learning}.
\newblock {\it \bibinfo{journal}{Transportation Research Part D: Transport and
  Environment}\/},  {\it \bibinfo{volume}{146}\/}, \bibinfo{pages}{104899}.
\bibitem[{Zhao et~al.(2024)Zhao, Hu and Zhang}]{zhao2024evaluating}
\bibinfo{author}{Zhao, Y.}, \bibinfo{author}{Hu, S.}, and
  \bibinfo{author}{Zhang, M.} (\bibinfo{year}{2024}).
\newblock \bibinfo{title}{Evaluating equitable transit-oriented development
  (tod) via the node-place-people model}.
\newblock {\it \bibinfo{journal}{Transportation Research Part A: Policy and
  Practice}\/},  {\it \bibinfo{volume}{185}\/}, \bibinfo{pages}{104116}.
\bibitem[{Zheng et~al.(2023{\natexlab{a}})Zheng, Lin, Zhao, Wu, Jin and
  Li}]{zheng2023spatial}
\bibinfo{author}{Zheng, Y.}, \bibinfo{author}{Lin, Y.}, \bibinfo{author}{Zhao,
  L.}, \bibinfo{author}{Wu, T.}, \bibinfo{author}{Jin, D.}, and
  \bibinfo{author}{Li, Y.} (\bibinfo{year}{2023}{\natexlab{a}}).
\newblock \bibinfo{title}{Spatial planning of urban communities via deep
  reinforcement learning}.
\newblock {\it \bibinfo{journal}{Nature Computational Science}\/},  {\it
  \bibinfo{volume}{3}\/}, \bibinfo{pages}{748--762}.
\bibitem[{Zheng et~al.(2023{\natexlab{b}})Zheng, Su, Ding, Jin and
  Li}]{zheng2023road}
\bibinfo{author}{Zheng, Y.}, \bibinfo{author}{Su, H.}, \bibinfo{author}{Ding,
  J.}, \bibinfo{author}{Jin, D.}, and \bibinfo{author}{Li, Y.}
  (\bibinfo{year}{2023}{\natexlab{b}}).
\newblock \bibinfo{title}{Road planning for slums via deep reinforcement
  learning}.
\newblock In {\it \bibinfo{booktitle}{Proceedings of the 29th ACM SIGKDD
  Conference on Knowledge Discovery and Data Mining}\/} (pp.
  \bibinfo{pages}{5695--5706}).
\bibitem[{Zhou et~al.(2020)Zhou, Yang and Webster}]{zhou2020using}
\bibinfo{author}{Zhou, J.}, \bibinfo{author}{Yang, Y.}, and
  \bibinfo{author}{Webster, C.} (\bibinfo{year}{2020}).
\newblock \bibinfo{title}{Using big and open data to analyze transit-oriented
  development: new outcomes and improved attributes}.
\newblock {\it \bibinfo{journal}{Journal of the American Planning
  Association}\/},  {\it \bibinfo{volume}{86}\/}, \bibinfo{pages}{364--376}.

\end{thebibliography}

\end{document}